%% file: main.tex
\documentclass[10pt,twocolumn,letterpaper]{article}

\usepackage[pagenumbers]{cvpr} % To force page numbers, e.g. for an arXiv version

\usepackage[accsupp]{axessibility}  % Improves PDF readability for those with disabilities.

\usepackage{graphicx}
\usepackage{amsmath}
\usepackage{amssymb}
\usepackage{booktabs}

\usepackage{amsmath}
\usepackage{amssymb}
\usepackage[table,xcdraw]{xcolor}
\usepackage{pifont}
\usepackage{multirow}
\usepackage{makecell}
\usepackage[normalem]{ulem}
\useunder{\uline}{\ul}{}
\usepackage{endnotes}

\usepackage[table]{xcolor}

\usepackage{colortbl}
\definecolor{colorh}{rgb}{1,0.60,0.20}
\definecolor{colorm}{rgb}{1,0.72,0.30}
\definecolor{colorl}{rgb}{1,0.88,0.70}
\newcommand{\colorh}[1]{\colorbox{colorh}{{#1}}}
\newcommand{\colorm}[1]{\colorbox{colorm}{{#1}}}
\newcommand{\colorl}[1]{\colorbox{colorl}{{#1}}}

\usepackage[pagebackref,breaklinks,colorlinks]{hyperref}

\usepackage[capitalize]{cleveref}
\crefname{section}{Sec.}{Secs.}
\Crefname{section}{Section}{Sections}
\Crefname{table}{Table}{Tables}
\crefname{table}{Tab.}{Tabs.}

\def\cvprPaperID{***} % *** Enter the CVPR Paper ID here
\def\confName{CVPR}
\def\confYear{2023}

\begin{document}

%%%%%%%%% TITLE - PLEASE UPDATE
\title{EDA: Explicit Text-Decoupling and Dense Alignment for 3D Visual Grounding}

\author{
    Yanmin Wu\textsuperscript{\rm 1} \quad ~
    Xinhua Cheng\textsuperscript{\rm 1} \quad ~
    Renrui Zhang\textsuperscript{\rm 2,3} \quad ~
    Zesen Cheng\textsuperscript{\rm 1} \quad ~
    Jian Zhang\textsuperscript{\rm 1}$^*$
    \vspace{5pt}
    \\
    \textsuperscript{\rm 1} Shenzhen Graduate School, Peking University, China \\
    \textsuperscript{\rm 2} The Chinese University of Hong Kong, China  ~%\quad
    %\textsuperscript{\rm 3} Peng Cheng Laboratory \quad
    \textsuperscript{\rm 3} Shanghai AI Laboratory, China
    \\
{\tt\small wuyanmin@stu.pku.edu.cn} \quad \quad
{\tt\small zhangjian.sz@pku.edu.cn}}
%\maketitle

\twocolumn[{%
\renewcommand\twocolumn[1][]{#1}%

\maketitle

\vspace{-20pt}
\begin{center}
    \centering
    \captionsetup{type=figure}
    \includegraphics[width=1.0 \textwidth]{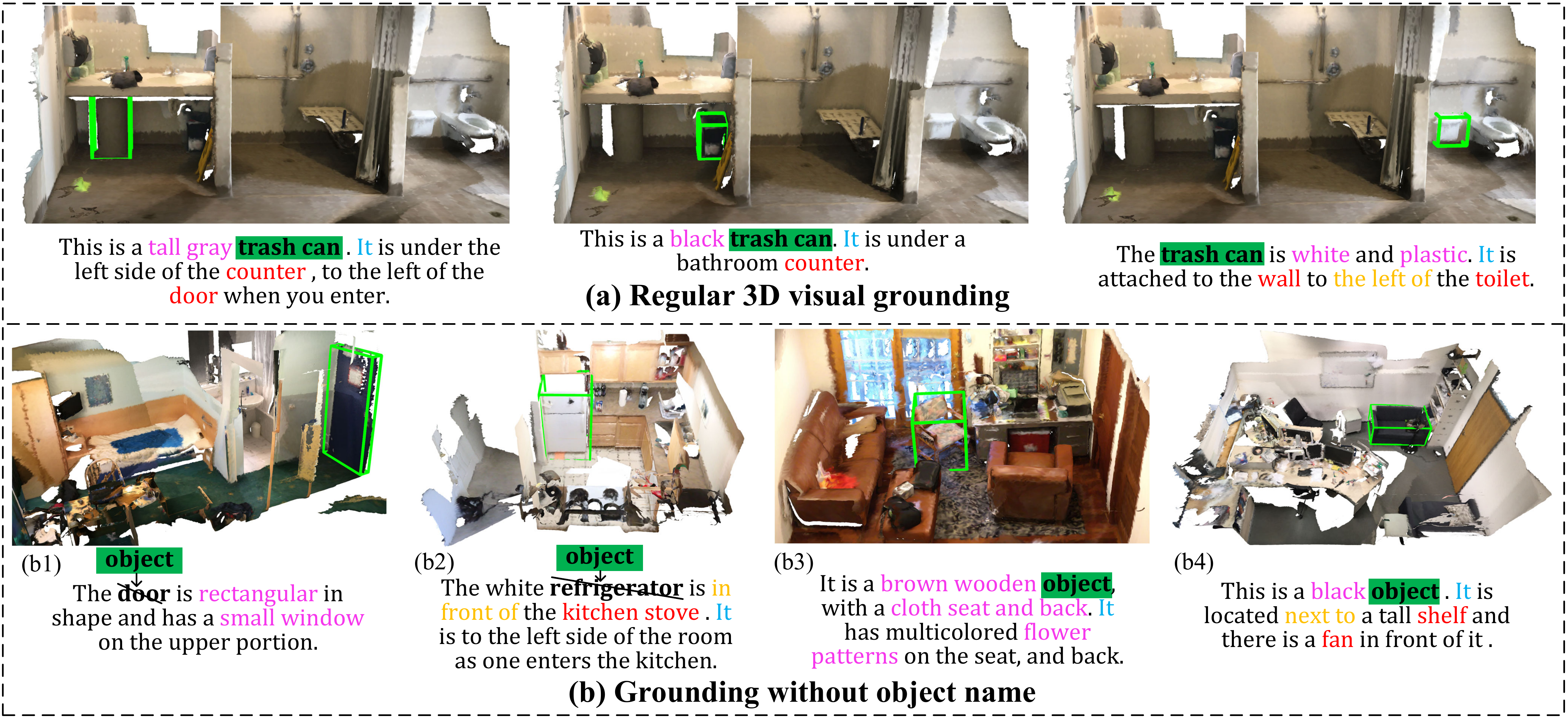}
    \captionof{figure}{Text-decoupled, dense aligned 3D visual grounding. Different colours in the text correspond to different decoupled components. (a) Regular 3D visual grounding: locating objects requires comprehensively considering multiple semantic cues such as appearance attributes, object names, and spatial relationships. (b) Grounding without object name: not mentioning object names, avoiding short-cuts and forcing the model to predict the target based on other attributes.}
    \label{fig:demo}
\end{center}
}]

\renewcommand{\thefootnote}{}
\footnotetext{$^*$\textit{Corresponding author}. This work was supported in part by Shenzhen Research Project under Grant JCYJ20220531093215035.}

%%%%%%%%%%%%%%%%%%%%%%%%%%%%%%%%%%%%%%%%%%%%%%%%%%%%%%%
% 0. abstract
%%%%%%%%%%%%%%%%%%%%%%%%%%%%%%%%%%%%%%%%%%%%%%%%%%%%%%%
\begin{abstract}
\vspace{-5pt}
3D visual grounding aims to find the object within point clouds mentioned by free-form natural language descriptions with rich semantic cues. However, existing methods either extract the sentence-level features coupling all words or focus more on object names, which would lose the word-level information or neglect other attributes.
To alleviate these issues, we present \textbf{EDA} that \textbf{E}xplicitly \textbf{D}ecouples the textual attributes in a sentence and conducts \textbf{D}ense \textbf{A}lignment between such fine-grained language and point cloud objects.
Specifically, we first propose a text decoupling module to produce textual features for every semantic component. Then, we design two losses to supervise the dense matching between two modalities: position alignment loss and semantic alignment loss.
On top of that, we further introduce a new visual grounding task, locating objects without object names, which can thoroughly evaluate the model's dense alignment capacity.
Through experiments, we achieve state-of-the-art performance on two widely-adopted 3D visual grounding datasets, ScanRefer and SR3D/NR3D, and obtain absolute leadership on our newly-proposed task. The source code is available at \url{https://github.com/yanmin-wu/EDA}.
%The source code and pre-trained models will be available at \url{https://github.com/yanmin-wu/EDA}.
\end{abstract}

\vspace{-10pt}
%%%%%%%%%%%%%%%%%%%%%%%%%%%%%%%%%%%%%%%%%%%%%%%%%%%%%%%
% 1. Introduction
%%%%%%%%%%%%%%%%%%%%%%%%%%%%%%%%%%%%%%%%%%%%%%%%%%%%%%%
\section{Introduction}
\label{sec:intro}

Multi-modal cues can highly benefit the 3D environment perception of an agent, including 2D images, 3D point clouds, and language. Recently, \textbf{3D visual grounding} (3D VG) \cite{chen2020scanrefer,Prabhudesai_2020_CVPR}, also known as 3D object referencing \cite{achlioptas2020referit3d}, has attached much attention as an important 3D cross-modal task. Its objective is to find the target object in point cloud scenes by analyzing the descriptive query language, which requires understanding both 3D visual and linguistic context.

Language utterances typically involve words describing appearance attributes, object categories, spatial relationships and other characteristics, as shown by different colours in Fig.~\ref{fig:demo}(a), requiring that the model integrate multiple cues to locate the mentioned object.
Compared with 2D Visual Grounding \cite{deng2018visual,deng2021transvg,yang2022improving,zhang2022glipv}, the sparseness and incompleteness of point clouds, and the diversity of language descriptions produced by 3D multi-view, make 3D VG more challenging.
Existing works made significant progress from the following perspectives: improving point cloud features extraction by sparse convolution \cite{yuan2021instancerefer} or 2D images assistance \cite{yang2021sat}; generating more discriminative object candidates through instance segmentation \cite{huang2021text} or language modulation \cite{luo20223d}; identifying complex spatial relationships between entities via graph convolution \cite{feng2021free} or attention \cite{cai20223djcg}.

However, we observe two issues that remain unexplored. 
\textbf{1) Imbalance}: The object name can exclude most candidates, and even in some cases, there is only one name-matched object, as the \textit{``door"} and \textit{``refrigerator"} in Fig.~\ref{fig:demo}(b1, b2). This shortcut may lead to an inductive bias in the model that pays more attention to object names while weakening other properties such as appearance and relationships, resulting in imbalanced learning.
\textbf{2) Ambiguity}: Utterances frequently refer to multiple objects and attributes (such as \textit{``black object, tall shelf, fan"} in Fig.~\ref{fig:demo}(b4)), while the model's objective is to identify only the main object, leading to an ambiguous understanding of language descriptions.
These insufficiencies of existing works stem from their characteristic of feature coupling and fusing implicitly. They input a sentence with different attribute words but output only one globally \textbf{coupled} sentence-level feature that subsequently matches the visual features of candidate objects. The coupled feature is ambiguous because some words may not describe the main object (green text in Fig.~\ref{fig:demo}) but other auxiliary objects (red text in Fig.~\ref{fig:demo}). 
Alternatively, using the cross-modal attention of the Transformer \cite{vaswani2017attention, dosovitskiy2021an} automatically and \textbf{implicitly} to fuse visual and text features. However, this may encourage the model to take shortcuts, such as focusing on object categories and ignoring other attributes, as previously discussed.

Instead, we propose a more intuitive decoupled and explicit strategy. First, we parse the input text to \textbf{decouple} different semantic components, including the main object word, pronoun, attributes, relations, and auxiliary object words. Then, performing \textbf{dense alignment} between point cloud objects and multiple related decoupled components achieves fine-grained feature matching, which avoids the inductive bias resulting from imbalanced learning of different textual components. As the final grounding result, we \textbf{explicitly} select the object with the highest similarity to the decoupled text components (instead of the entire sentence), avoiding ambiguity caused by irrelevant components.
Additionally, to explore the limits of VG and examine the comprehensiveness and fine-graininess of visual-language perception of the model, we suggest a challenging new task: \textbf{Grounding without object name (VG-w/o-ON)}, where the name is replaced by \textit{``object"} (see Fig.~\ref{fig:demo}(b)), forcing the model to locate objects based on other attributes and relationships. This setting makes sense because utterances that do not mention object names are common expressions in daily life, and in addition to testing whether the model takes shortcuts.
Benefiting from our text decoupling operation and the supervision of dense aligned losses, all text components are aligned with visual features, making it possible to locate objects independent of object names.

To sum up, the main contributions of this paper are as follows: 
\textbf{1)} We propose a text decoupling module to parse linguistic descriptions into multiple semantic components, followed by suggesting two well-designed dense aligned losses for supervising fine-grained visual-language feature fusion and preventing imbalance and ambiguity learning. 
\textbf{2)} The challenging new 3D VG task of grounding without object names is proposed to comprehensively examine the model's robust performance.
\textbf{3)} We achieve state-of-the-art performance on two datasets (ScanRefer and SR3D/NR3D) on the regular 3D VG task and absolute leadership on the new task evaluated by the same model without retraining.

%%%%%%%%%%%%%%%%%%%%%%%%%%%%%%%%%%%%%%%%%%%%%%%%%%%%%%%
% 2. Related Work
%%%%%%%%%%%%%%%%%%%%%%%%%%%%%%%%%%%%%%%%%%%%%%%%%%%%%%%
\section{Related Work}

%%%%%%%%% 2.1 %%%%%%%%%%%%%
\subsection{3D Vision and Language}
3D vision~\cite{qi2017pointnet++,qi2017pointnet,zhang2023parameter} and language are vital manners for humans to understand the environment, and they are also important research topics for the evolution of machines to be like humans. 
Previously, the two fields evolved independently. Due to the advance of multimodality \cite{radford2021learning, ramesh2022hierarchical, qi2020reverie, zhang2022pointclip,guo2023joint,guo2022calip,zhang2022learning,zhang2022can,zhang2023prompt}, many promising works across 3D vision and language have been introduced recently.
In 3D visual grounding \cite{chen2020scanrefer,chen2021d3net,chen2022ham,abdelreheem2022scanents3d,chen2022language}, the speaker (like a human) describes an object in language. The listener (such a robot) needs to understand the language description and the 3D visual scene to grounding the target object. 
On the contrary, the 3D dense caption \cite{chen2021scan2cap, chen2021d3net, yuan2022x, jiao2022more, wang2022spatiality, chen2023end} is analogous to an inverse process in which the input is a 3D scene, and the output is textual descriptions of each object. 
Language-modulated 3D detection or segmentation \cite{jain2022bottom, rozenberszki2022language,chen2023pimae,huang2022tig,zhu2022pointclip} enriches the diversity of text queries by matching visual-linguistic feature spaces rather than predicting the probability of a set number of categories.
Furthermore, some studies explore the application of 3D visual language in agents such as robot perception \cite{ha2022semantic, shafiullah2022clip}, vision-and-language navigation (VLN) \cite{hong2021vln, qiao2022hop, cheng2022learning}, and embodied question answering (EQA) \cite{tan2021knowledge, azuma2022scanqa, luo2022depth, ma2022sqa3d, etesam20223dvqa, hong2023threedclr}.
In this paper, we focus on point clouds-based 3D visual grounding, which is the fundamental technology for many embodied AI \cite{khandelwal2022simple, paul2022avlen,gadre2022clip} tasks.

%%%%%%%%% 2.2 %%%%%%%%%%%%%
\subsection{3D Visual Grounding}
The majority of current mainstream techniques are two-stage. In the first stage, obtain the features of the query language and candidate point cloud objects independently by a pre-trained language model~\cite{pennington2014glove,devlin2018bert,chung2014empirical} and a pre-trained 3D detector~\cite{qi2019deep,liu2021group} or segmenter~\cite{jiang2020pointgroup,chen2021hierarchical,vu2022softgroup}. In the second stage, the researchers focus on fusing the two modal features and then selecting the best-matched object.
\textbf{1)} 
The most straightforward solution is to concatenate the two modal features and then consider it a binary classification problem~\cite{chen2020scanrefer, kumar2022auto}, which provides limited performance because the two features are not sufficiently fused.
\textbf{2)} 
Taking advantage of the Transformer's attention mechanism, which is naturally suitable for multi-module feature fusion, He \textit{et al.}~\cite{he2021transrefer3d} and Zhao \textit{et al.}~\cite{zhao20213dvg} achieve remarkable performance by performing self-attention and cross-attention to features.
\textbf{3)} In contrast, other studies view feature fusion as a matching problem rather than a classification. Yuan \textit{et al.}~\cite{yuan2021instancerefer} and Abdelreheem \textit{et al.}~\cite{abdelreheem20223dreftransformer}, supervised by the contrastive loss \cite{he2020momentum}, compute the cosine similarity of visual features and textual features. Inspired by \cite{liu2020learning}, Feng \textit{et al.}~\cite{feng2021free} parses the text to generate a text scene graph, simultaneously builds a visual scene graph, and then performs graph node matching.
\textbf{4)} Point clouds' sparse, noisy, incomplete, and lack of detail make learning objects' semantic information challenging. Yang \textit{et al.}~\cite{yang2021sat} and Cai \textit{et al.}~\cite{cai20223djcg} use 2D images to aid visual-textual feature fusion, but at the cost of additional 2D-3D alignment and 2D feature extraction.

However, the two-stage method has a substantial detection bottleneck: objects overlooked in the first stage cannot be matched in the second.
In contrast, object detection and feature extraction in the single-stage method is modulated by the query text, making it easier to identify the text-concerned object.
Liu \textit{et al.}~\cite{liu2021refer} suggest fusing visual and linguistic features at the bottom level and producing text-related visual heatmaps. Similarly, Luo \textit{et al.}~ \cite{luo20223d} present a single-stage approach that employs textual features to guide visual keypoint selection and progressively localizes objects.
BUTD-DETR~\cite{jain2022bottom} is also a single-stage capable framework. More importantly, inspired by the 2D image-language pre-train model (such MDETR~\cite{kamath2021mdetr}, GLIP~\cite{li2022grounded}), BUTD-DETR measures the similarity between each word and object and then selects the features of the word that correspond to the object's name to match the candidate object. 
However, there are two limitations: 1) Since multiple object names may be mentioned in a sentence, the ground truth annotation is needed to retrieve the target name, which limits its generalizability. Our text decoupling module separates text components and determines the target object name by grammatical analysis to avoid this restriction. 2) BUTD-DETR (and MDETR and GLIP in the 2D task) only consider the \textbf{sparse alignment} of main object words or noun phrases to visual features. Conversely, we align all object-related decoupled textual semantic components with visual features, which we refer to \textbf{dense alignment}, significantly enhancing the discriminability of multimodal features.

\begin{figure}[!t]
\centering
\includegraphics[width=0.45 \textwidth]{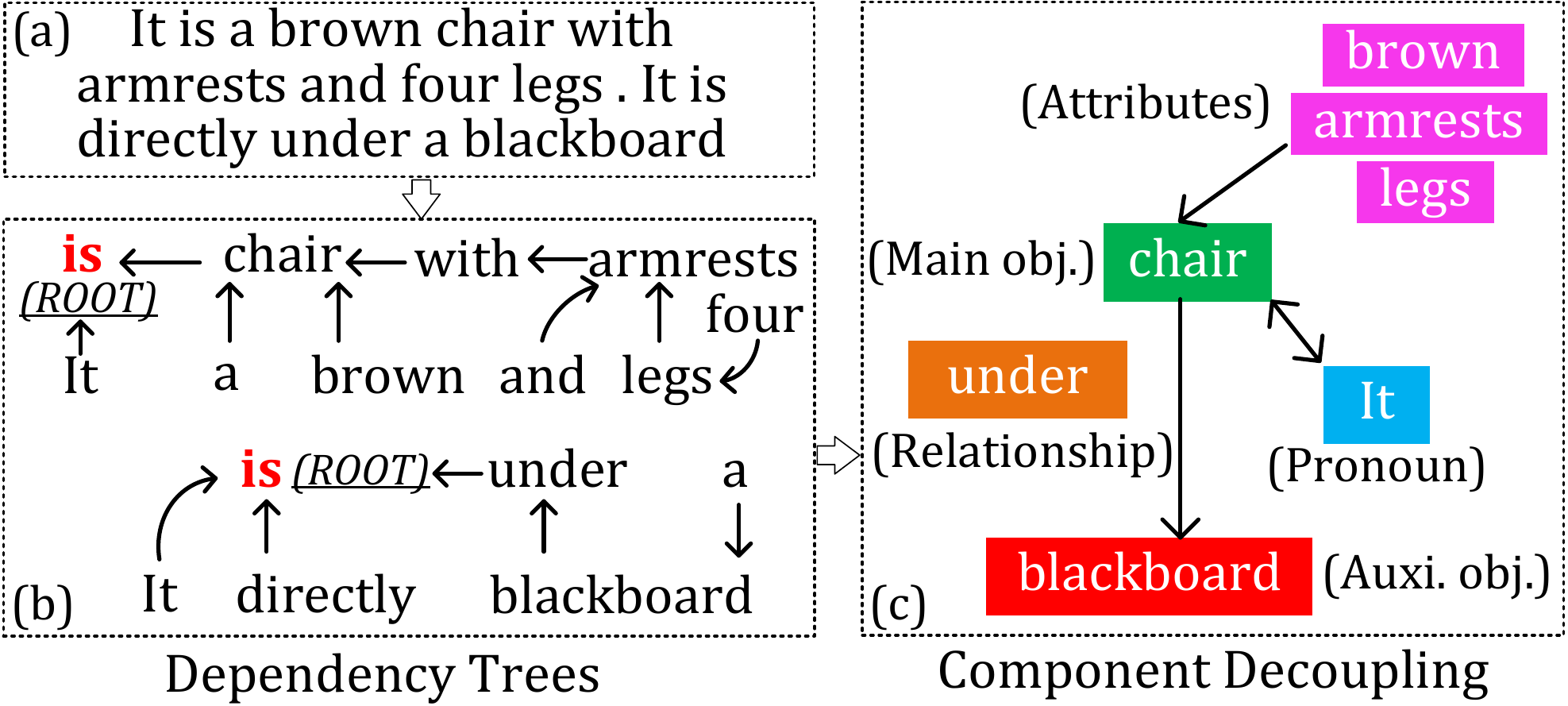} 
\caption{Text component decoupling: (a) The query text. (b) Dependency tree analysis. (c) Decoupled into five components.}
\label{fig:Text_Decouping}
\vspace{-8pt}
\end{figure}

\begin{figure*}[!t]
\centering
\includegraphics[width=1.0 \textwidth]{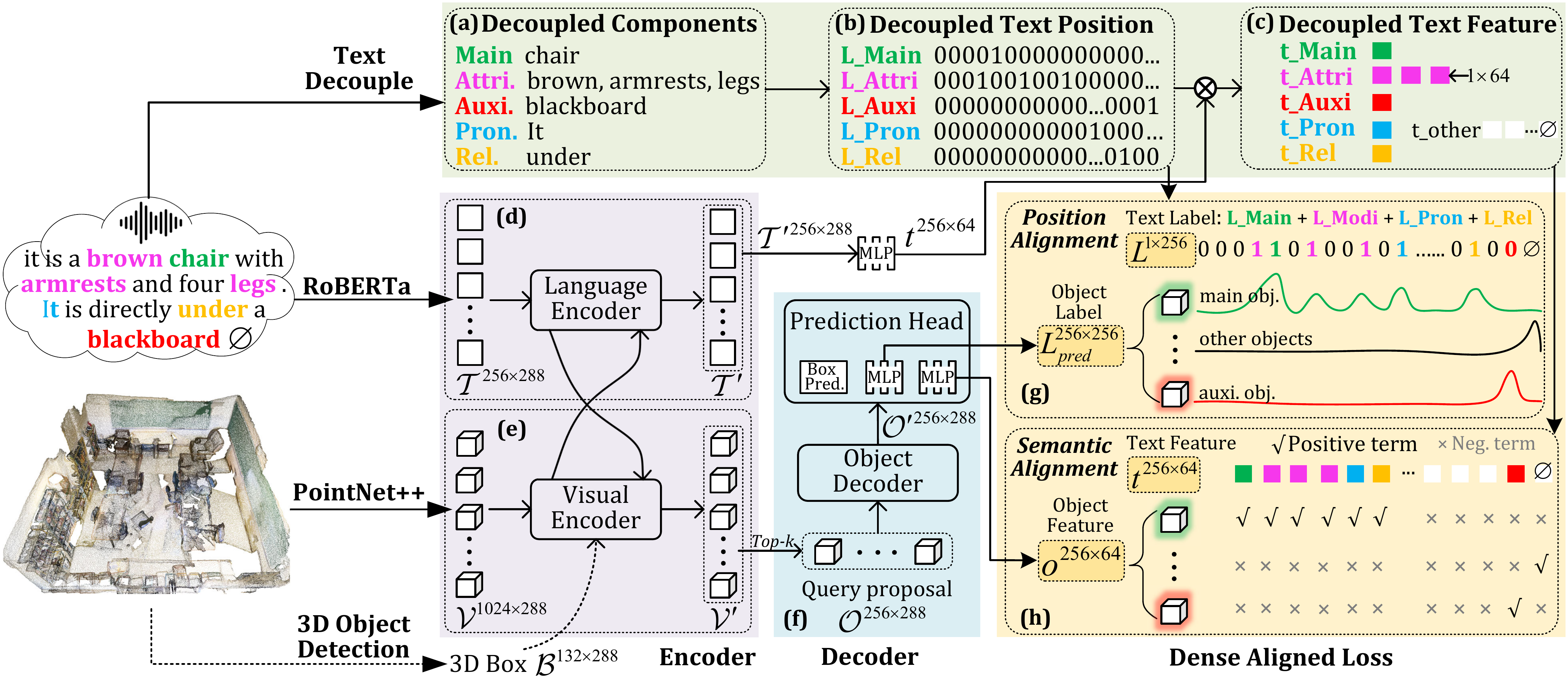} 
\caption{The system framework. (a-c): Decouple the input text into several components to acquire the position label $L$ and features $\boldsymbol{t}$ of the decoupled text. (d-e): Transformer-based encoders for cross-modal visual-text feature extraction. 
(f): Decode proposal features $\mathcal{O'}$ and linearly project them as object position labels $L_{pred}$ and object features $\boldsymbol{o}$, in addition to a box prediction head for regression of the bounding box.
(g-h): Visual-text feature dense alignment. Note that the additional 3D object detection procedure is optional.
%without it, the method degenerates into a single-stage fashion in which object detection and feature extraction are carried out in (e).
}
\label{fig:Framework}
\end{figure*}

%%%%%%%%%%%%%%%%%%%%%%%%%%%%%%%%%%%%%%%%%%%%%%%%%%%%%%%
% 3. Proposed Method
%%%%%%%%%%%%%%%%%%%%%%%%%%%%%%%%%%%%%%%%%%%%%%%%%%%%%%%
\section{Proposed Method}

The framework is illustrated in Fig.~\ref{fig:Framework}. First, the input text description is decoupled into multiple semantic components, and its affiliated text positions and features are obtained (Sec. \ref{sec:text}). Concurrently, the Transformer-based encoder extracts and modulates features from point clouds and text, then decodes the visual features of candidate objects (Sec. \ref{sec:backbone}).
Finally, the dense aligned losses are derived between the decoupled text features and the decoded visual features (Sec. \ref{sec:loss}). The grounding result is the object with visual features most similar to text features (Sec. \ref{sec:score}).

%%%%%%%%% 3.1 %%%%%%%%%%%%%
\subsection{Text Decoupling}
\label{sec:text}
The text features of the coupled strategy are ambiguous, where features from multiple objects and attributes are coupled, such as \textit{``a brown wooden chair next to the black table."} Among them, easy-to-learn clues (such as the category \textit{``chair "} or the colour \textit{``brown"}) may predominate, weakening other attributes (such as material \textit{``wooden"}); words of other objects (such as the \textit{``black table"}) may cause interference. To produce more discriminative text features and fine-grained cross-modal feature fusion, we decouple the query text into different semantic components, each independently aligned with visual features, avoiding the ambiguity caused by feature coupling.

\textbf{Text Component Decoupling.} 
Analyzing grammatical dependencies between words is a fundamental task in NLP. We first use the off-the-shelf tool \cite{schuster2015generating, wu2019unified} to parse the language description grammatically to generate the grammatical dependency trees, as shown in Fig.~\ref{fig:Text_Decouping}(b). Each sentence contains only one ROOT node, and each remaining word has a corresponding parent node. Then according to the words' part-of-speech and dependencies, we decouple the long text into \textbf{\textit{five semantic components}} (see Fig.~\ref{fig:Text_Decouping}(c)): \texttt{Main object} - the target object mentioned in the utterance; \texttt{Auxiliary object} - the one used to assist in locating the main object; \texttt{Attributes} - objects' appearance, shape, \textit{etc.}; \texttt{Pronoun} - the word instead of the main object; \texttt{Relationship} - the spatial relation between the main object and the auxiliary object. Note that attributes affiliated with pronouns are equivalent to attached with the main object, thus connecting two sentences in an utterance.

\textbf{Text Position Decoupling.} 
After decoupling each text component (Fig.~\ref{fig:Framework}(a)), we generate the position label (similar to a mask) $L_{main}, L_{attri}, L_{auxi}, L_{pron}, L_{rel} \in \mathbb{R}^{1 \times l}$ for the component's associated word (Fig.~\ref{fig:Framework}(b)). Where $l$$=$$256$ is the maximum length of the text, each component's word position is set to 1 and the rest to 0. The label will be used to construct the position alignment loss and supervise the classification of objects. The classification result, is not one of a predetermined number of object categories but the position of the text with the highest semantic similarity.

\textbf{Text Feature Decoupling.} 
The feature of each word (token) is produced in the backbone of multimodal feature extraction (Fig.~\ref{fig:Framework}(d)). The text feature of the decoupled component can be derived by dot-multiplying all words' features $\boldsymbol{t}$ with its position label $L$, as shown in Fig.~\ref{fig:Framework}(c). The decoupled text features and visual features will be independently aligned under the supervision of the semantic alignment loss.
Note that in the decoupled text features, the semantics of the corresponding components absolutely predominate, but as a result of the Transformer's attention mechanism, it also implicitly contains the global sentence's information. In other words, feature decoupling produces individual features while keeping the global context.

%%%%%%%%% 3.2 %%%%%%%%%%%%%
\subsection{Multimodal Feature Extraction}
\label{sec:backbone}
We employ BUTD-DETR's encoder-decoder module for feature extraction and intermodulation of cross-modal features. We strongly recommend the reader to refer to Fig.~\ref{fig:Framework}.
%as shown in Fig.~\ref{fig:Framework}(d)(e)(f). 

\textbf{Input Modal Tokenlization.} 
The input text and 3D point clouds are encoded by the pre-trained RoBERTa~\cite{liu2019roberta} and PointNet++~\cite{qi2017pointnet++} and produce text tokens $\mathcal{T} \in \mathbb{R}^{l \times d}$ and visual tokens $\mathcal{V} \in \mathbb{R}^{n \times d}$.
Additionally, the GroupFree~\cite{liu2021group} detector is used to detect 3D boxes, which are subsequently encoded as box tokens $\mathcal{B} \in \mathbb{R}^{b \times d}$.
Note that the GroupFree is \textbf{\textit{optional}}, the final predicted object of the network is from the prediction head (see below), and the box token is just to assist in better regression of the target object. 

\textbf{Encoder-Decoder.} 
% encoder
Self-attention and cross-attention are performed in the encoder to update both visual and text features, obtaining cross-modal features $\mathcal{V'}, \mathcal{T'}$ while keeping the dimensions.
% decoder
The top-$k$ ($k$$=$$256$) visual features are selected, linearly projected as query proposal features $\mathcal{O} \in \mathbb{R}^{k \times d}$, and updated as $\mathcal{O'}$ in the decoder.

\textbf{Prediction Head.}
1) The decoded proposal features $\mathcal{O'} \in \mathbb{R}^{k \times d}$ are fed into an MLP and output the predicted position labels $L_{pred} \in \mathbb{R}^{k \times l}$, which are then utilized to calculate the position alignment loss with decoupled text position labels $L \in \mathbb{R}^{1 \times l}$. 
2) Additionally, the proposal features are linearly projected as object features $\boldsymbol{o} \in \mathbb{R}^{k \times 64}$ by another MLP, which are then utilized to compute the semantic alignment loss with the similarly linearly projected text feature $\boldsymbol{t} \in \mathbb{R}^{l \times 64}$.
3) Lastly, a box prediction head~\cite{liu2021group} regresses the bounding box of the object.

%%%%%%%%% 3.3 %%%%%%%%%%%%%
\subsection{Dense Aligned Loss}
\label{sec:loss}

\subsubsection{Dense Position Aligned Loss} 
The objective of position alignment is to ensure that the distribution of language-modulated visual features closely matches that of the query text description, as shown in Fig.~\ref{fig:Framework}(g).
This process is similar to standard object detection's one-hot label prediction. However, rather than being limited by the number of categories, we predict the position of text that is similar to objects.

The constructed ground truth \textit{text distribution} of the mentioned main object is obtained by element-wise summing the position labels of the associated decoupled text components:
\begin{equation}
    P_{main} = \lambda_1 L_{main} + \lambda_2 L_{attri} + \lambda_3 L_{pron} + \lambda_4 L_{rel},
    \label{eq:P_text}
\end{equation}
where $\lambda$ is the weight of different parts (refer to the parametric search in \textcolor{magenta}{Supplementary Material}.). $P_{auxi} = L_{auxi}$ represents the text distribution of the auxiliary object. The remaining candidate objects' text distribution is $P_{oth}$, with the final bit set to 1 (see $\varnothing$ in Fig.~\ref{fig:Framework}(g)). Therefore, all $k$ candidate objects' ground truth text distribution is $P_{text} = \left \{ P_{main},P_{auxi},P_{oth} \right \} \in \mathbb{R}^{k \times l}$.

The predicted \textit{visual distribution} of $k$ objects is produced by applying softmax to the output $L_{pred} \in \mathbb{R}^{k \times l}$ of the prediction head:
\begin{equation}
    P_{obj} = Softmax(L_{pred}).
    \label{eq:P_obj}
\end{equation}
Their KL divergence is defined as the position-aligned loss:
\begin{equation}
    \mathcal{L}_{pos} =  \sum_{i=1}^{k}[P_{text}^{i} \log({P_{text}^{i}}) - P_{text}^{i} \log(P_{obj}^{i}) ].
    \label{eq:loss_possition}
\end{equation}

We \textbf{highlight} that ``dense alignment" indicates that the target object is aligned with the positions of multiple components (Eq.~(\ref{eq:P_text})), significantly different from BUTD-DETR (and MDETR for 2D tasks), which only sparsely aligns with the object name's position $L_{main}$.

\subsubsection{Dense Semantic Aligned Loss} 
Semantic alignment aims to learn the similarity of visual-text multimodal features through contrastive learning. The object loss of semantic alignment is defined as follows: 
\begin{small}
\begin{equation}
    \mathcal{L}_{sem\_o}\! =\! \sum_{i=1}^{k} \frac{1}{\left|\mathbf{T}_{i}^{+}\right|} \!\sum_{\boldsymbol{t}_i \in \mathbf{T}_{i}^{+}}\!\!\!-\log \!\left(\!\frac{\exp \left(w_{+}\ast (\boldsymbol{o}_{i}^{\top} \boldsymbol{t}_{i} / \tau)\right)}{\sum_{j=1}^{l} \exp \left(w_{-}\ast (\boldsymbol{o}_{i}^{\top} \boldsymbol{t}_{j} / \tau)\right)}\!\right),
    \label{eq:loss_semantic}
\end{equation}
\end{small}
where $\boldsymbol{o}$ and $\boldsymbol{t}$ are the object and text features after linear projection, and $\boldsymbol{o}^{\top} \boldsymbol{t} / \tau$ is their similarity, as shown in Fig.~\ref{fig:Framework} (h).
$k$ and $l$ are the number of objects and words.
$\boldsymbol{t}_i$ is the positive text feature of the $i_{th}$ candidate object. Taking the main object as an example, the positive text feature $\mathbf{T}_i^{+}$ corresponding to it is:
\begin{equation}
    \boldsymbol{t}_i \in \mathbf{T}_i^{+} = \left \{ \boldsymbol{t}_{main}, \boldsymbol{t}_{attri}, \boldsymbol{t}_{pron}, \boldsymbol{t}_{rel} \right \},
    \label{eq:text_positive}
\end{equation}
and $w_{+}$ is the weight of each positive term. $\boldsymbol{t}_j$ is the feature of the $i_{th}$ text, but note that the negative similarity weight $w_{-}$ for auxiliary object term $\boldsymbol{t}_{auxi}$ is 2, while the rest weight 1.
The text loss of semantic alignment defined similarly:
\begin{small}
\begin{equation}
    \mathcal{L}_{sem\_t} = \sum_{i=1}^{l} \frac{w_{+}}{\left|\mathbf{O}_{i}^{+}\right|} \sum_{\boldsymbol{o}_i \in \mathbf{O}_{i}^{+}}-\log \left(\frac{\exp \left (\boldsymbol{t}_{i}^{\top} \boldsymbol{o}_{i} / \tau\right)}{\sum_{j=1}^{k} \exp \left (\boldsymbol{t}_{i}^{\top} \boldsymbol{o}_{j} / \tau \right)}\right),
    \label{eq:loss_semantic_text}
\end{equation}
\end{small}
where $\boldsymbol{o}_i \in \mathbf{O}_{i}^{+}$ is the positive object feature of the $i_{th}$ text, and $\boldsymbol{o}_{j}$ is the feature of the $j_{th}$ object.
The final semantic alignment loss is the mean of the two: $\mathcal{L}_{sem} = (\mathcal{L}_{sem\_o} + \mathcal{L}_{sem\_t})/2.$

Similarly, the semantic alignment of multiple text components (Eq.~(\ref{eq:text_positive})) with visual features also illustrates our insight of ``dense." This is intuitive, such as \textit{``It is a brown chair with legs under a blackboard,"} where the main object's visual features should be not only similar to \textit{``chair"} but also similar to \textit{``brown, legs"} and distinct to \textit{``blackboard"} as possible. 

The total loss for training also includes the box regression loss. Refer to \textcolor{magenta}{Supplementary Material} for details.

%%%%%%%%% 3.4 %%%%%%%%%%%%%
\subsection{Explicit Inference}
\label{sec:score}

Because of our text decoupling and dense alignment operations, object features fused multiple related text component features, allowing for the computation of the similarity between individual text components and candidate objects. For instance, $S_{main}=Softmax(\boldsymbol{o}^{\top} \boldsymbol{t}_{main} / \tau)$ indicates the similarity between objects $\boldsymbol{o}$ and the main text component $\boldsymbol{t}_{main}$. Therefore, the similarity of objects and related components can be explicitly combined to obtain the total score and select the candidate with the highest score:
\begin{equation}
    S_{all} = S_{main} + S_{attri} + S_{pron} + S_{rel} - S_{auxi},
    \label{eq:score}
\end{equation}
where the definition of $S_{attri}, S_{pron}, S_{rel}$, and $S_{auxi}$ is similar to $S_{main}$. If providing supervision of auxiliary objects during training, and the auxiliary object can be identified by solely computing the similarity between the object features and the auxiliary component's text features: $S_{all} = S_{attri}$.
Being able to infer the object based on the part of the text is a significant sign that the network has learned well-aligned and fine-grained visual-text feature space.

%%%%%%%%%%%%%%%%%%%%%%%%%%%%%%%%%%%%%%%%%%%%%%%%%%%%%%%
% 4. Experiments
%%%%%%%%%%%%%%%%%%%%%%%%%%%%%%%%%%%%%%%%%%%%%%%%%%%%%%%
\section{Experiments}
First, we conduct comprehensive and fair comparisons with SOTA methods in the Regular 3D Visual Grounding setting in Sec.~\ref{sec:RVG}. Then, in Sec.~\ref{sec:VG-w/o-ON}, we introduce our proposed new task, Grounding without Object Name, and perform comparison and analysis. Implementation details, additional experiments and more qualitative results are detailed in the \textcolor{magenta}{supplementary material}.

\begin{table*}[!t]
\small
\centering
\setlength{\tabcolsep}{5pt}
\renewcommand{\arraystretch}{1.0}
\setlength{\abovecaptionskip}{2pt}
\begin{tabular}{ccc|cc|cc|cc}
\toprule
\rowcolor[HTML]{E6E6E6}            &  &  & \multicolumn{2}{c|}{Unique ($\sim$19\%)} & \multicolumn{2}{c|}{Multiple ($\sim$81\%)} & \multicolumn{2}{c}{\textbf{Overall}}            \\
\rowcolor[HTML]{E6E6E6} \multirow{-2}{*}{Method}                       &  \multirow{-2}{*}{Venue}        &      \multirow{-2}{*}{Modality}                     & 0.25              & 0.5              & 0.25               & 0.5               & \textbf{0.25}          & \textbf{0.5}           \\ 
\midrule
\multirow{2}{*}{ScanRefer~\cite{chen2020scanrefer}}        & \multirow{2}{*}{ECCV2020} & 3D                        & 67.64             & 46.19            & 32.06              & 21.26             & 38.97                  & 26.10                   \\
                                  &                                                            & 3D+2D                     & 76.33             & 53.51            & 32.73              & 21.11             & 41.19                  & 27.40                   \\ 
\rowcolor[HTML]{F5F5F5} ReferIt3D~\cite{achlioptas2020referit3d}                         & ECCV2020            & 3D                        & 53.8              & 37.5             & 21.0               & 12.8              & 26.4                   & 16.9                   \\ 
TGNN~\cite{huang2021text}                              & AAAI2021                      & 3D                        & 68.61             & 56.80             & 29.84              & 23.18             & 37.37                  & 29.70                   \\ 
\rowcolor[HTML]{F5F5F5} InstanceRefer~\cite{yuan2021instancerefer}                     & ICCV2021              & 3D                        & 77.45             & 66.83            & 31.27              & 24.77             & 40.23                  & 32.93                  \\ 
SAT~\cite{yang2021sat}                               & ICCV2021                        & 3D+2D                     & 73.21             & 50.83            & 37.64              & 25.16             & 44.54                  & 30.14                  \\ 
\rowcolor[HTML]{F5F5F5} FFL-3DOG~\cite{feng2021free}                          & ICCV2021                       & 3D                        & 78.80              &  \colorl{67.94}       & 35.19              & 25.70              & 41.33                  & 34.01                  \\ 
\multirow{2}{*}{3DVG-Transformer~\cite{zhao20213dvg}} & \multirow{2}{*}{ICCV2021}      & 3D                        & 77.16             & 58.47            & 38.38              & 28.70              & 45.90                   & 34.47                  \\
                                  &                                                            & 3D+2D                     & 81.93             & 60.64            & 39.30               & 28.42             & 47.57                  & 34.67                  \\ 
\rowcolor[HTML]{F5F5F5} 3D-SPS~\cite{luo20223d}           & CVPR2022               & 3D+2D                     & \colorm{84.12}       &  66.72       & 40.32              & 29.82             & 48.82                  & 36.98                  \\ 
\multirow{2}{*}{3DJCG~\cite{cai20223djcg}}            & \multirow{2}{*}{CVPR2022}      & 3D                        & 78.75             & 61.30             & 40.13              & 30.08             & 47.62                  & 36.14                  \\
                                  &                                                            & 3D+2D                     & \colorl{83.47}        & 64.34            &  \colorl{41.39}         & \colorl{30.82}        &  \colorl{49.56}             & 37.33             \\ 
\rowcolor[HTML]{F5F5F5} BUTD-DETR~\cite{jain2022bottom} \dag                       & ECCV2022                    & 3D                        & 82.88              & 64.98             & \colorm{44.73}         & \colorm{33.97}        & \colorm{50.42}             & \colorm{38.60}             \\ 
D3Net~\cite{chen2021d3net}                        & ECCV2022                    & 3D+2D                        & -              & \colorh{\textbf{70.35}}             & -         & 30.50        & -             & \colorl{37.87}             \\ 
\rowcolor[HTML]{F5F5F5} \textbf{EDA}                     & -                                                          & 3D                        & \colorh{\textbf{85.76}}    & \colorm{68.57}   & \colorh{\textbf{49.13}}     & \colorh{\textbf{37.64}}    & \colorh{\textbf{54.59 \tiny{(+4.2\%)}}} & \colorh{\textbf{42.26 \tiny{(+3.7\%)}}} \\ 
\midrule
3D-SPS(single-stage)~\cite{luo20223d}           & CVPR2022         & 3D                        & 81.63             & 64.77            & 39.48              & 29.61             & 47.65                  & 36.43                  \\
BUTD-DETR (single-stage)~\cite{jain2022bottom}\ddag                      & ECCV2022                                                          & 3D                        & 81.47    & 61.24   & 44.20     & 32.81    & 49.76 & 37.05 \\
\textbf{EDA (single-stage)} \S                      & -                                                          & 3D                        & \textbf{86.40}    & \textbf{69.42}   & \textbf{48.11}     & \textbf{36.82}    & \textbf{53.83} & \textbf{41.70} \\
\bottomrule
\end{tabular}
\caption{The 3D visual grounding results on ScanRefer, accuracy evaluated by IoU 0.25 and IoU 0.5.
\dag~The accuracy is reevaluated using our parsed text labels because the performance reported by BUTD-DETR used ground truth text labels and ignored some challenging samples (see \textcolor{magenta}{supplementary materials} for more details). \S~Our single-stage implementation without the assistance of the additional 3D object detection step (dotted arrows in Fig.~\ref{fig:Framework}). \ddag~BUTD-DETR did not provide single-stage results and we retrained the model.} %We color code \colorh{best}, \colorm{second best} and \colorl{third best} performances.}
\label{tab:ScanRefer_result}
\vspace*{-1\baselineskip}
\end{table*}

%%%%%%%%% 4.1 %%%%%%%%%%%%%
\subsection{Regular 3D Visual Grounding}
\label{sec:RVG}
\subsubsection{Experiment settings}
We keep the same settings as existing works, with ScanRefer \cite{chen2020scanrefer} and SR3D/NR3D \cite{achlioptas2020referit3d} as datasets and Acc@0.25IoU and Acc@0.5IoU as metrics. Based on the visual data of ScanNet \cite{dai2017scannet}, \textbf{ScanRefer} adds 51,583 manually annotated text descriptions about objects. These complex and free-form descriptions involve object categories and attributes such as colour, shape, size, and spatial relationships. \textbf{SR3D/NR3D} is also proposed based on ScanNet, with SR3D including 83,572 simple machine-generated descriptions and NR3D containing 41,503 descriptions similar to ScanRefer's human annotation. 
The \textbf{\textit{difference}} is that in the ScanRefer configuration, detecting and matching objects are required, while SR3D/NR3D is simpler. It supplies GT boxes for all candidate objects and only needs to classify the classes of the boxes and choose the target object.

\vspace{-5pt}
\subsubsection{Comparison to the state of the art}

% scanrefer
~~~\textbf{ScanRefer.}
Table~\ref{tab:ScanRefer_result} reports the results on the ScanRefer dataset. 
\textbf{\romannumeral1)} Our method achieves state-of-the-art performance by a substantial margin, with an overall improvement of 4.2\% and 3.7\% to 54.59\% and 42.26\%.
\textbf{\romannumeral2)} Some studies \cite{chen2020scanrefer,yang2021sat,zhao20213dvg,luo20223d,cai20223djcg,chen2021d3net} proved that supplemented 2D images with detailed and dense semantics could learn better point cloud features. Surprisingly, we only use sparse 3D point cloud features and even outperformed 2D assistance methods. This superiority illustrates that our decoupling and dense alignment strategies mine more efficient and meaningful visual-text co-representations.
\textbf{\romannumeral3)} Another finding is that the accuracy of most existing techniques is less than 40\% and 30\% in the \textbf{``multiple''} setting because multiple means that the category of the target object mentioned in the language is not unique, with more interference candidates with the same category. However, we reached a remarkable 49.13\% and 37.64\%. To identify similar objects, a finer-grained understanding of the text and vision is required in this complex setting.
\textbf{\romannumeral4)} The last three rows in Table~\ref{tab:ScanRefer_result} compare single-stage methods, where our method's single-stage implementation is without the object detection step ($\mathcal{B}$ in Fig.~\ref{fig:Framework}) in training and inference. The result illustrates that while not requiring an additional pre-trained 3D object detector, our approach also can achieve SOTA performance. 
\textbf{\romannumeral5)} The qualitative results are depicted in Fig.~\ref{fig:vis}(a-c), which reveals that our method with an excellent perception of appearance attributes, spatial relationships, and even ordinal numbers.

\textbf{SR3D/NR3D.}
Table~\ref{tab:SR3D_result} shows the accuracy on the SR3D/NR3D dataset, where we achieve the best performance of 68.1\% and 52.1\%.
In SR3D, since the language descriptions are concise and the object is easy to identify, our method and \cite{luo20223d,jain2022bottom} reach an accuracy of over 60\%.
Conversely, in NR3D, descriptions are too detailed and complex, causing additional challenges for text decoupling. However, we still achieve SOTA accuracy with the 3D-only data, while other comparable methods~\cite{yang2021sat,luo20223d} rely on additional 2D images for training.
Some methods \cite{chen2020scanrefer,chen2021d3net,cai20223djcg} compared in Table~\ref{tab:ScanRefer_result} are not discussed here because they are not evaluated on the SR3D/NR3D dataset.
In addition, because GT boxes of candidate objects are provided in this setting, the single-stage methods are not applicable and discussed.

\begin{table}[!t]
\small
\centering
\setlength{\tabcolsep}{4.5pt}
\setlength{\abovecaptionskip}{2pt}
\begin{tabular}{ccc|cc}
\toprule
\rowcolor[HTML]{E6E6E6} Method                        & Venue        & Modality                            & SR3D          & NR3D          \\
\midrule
ReferIt3D \footnotesize{\cite{achlioptas2020referit3d}}     & ECCV20      & 3D                  & 39.8          & 35.6          \\
TGNN \footnotesize{\cite{huang2021text}}        & AAAI21      &3D                                 & 45.0            & 37.3          \\
TransRefer3D \footnotesize{\cite{he2021transrefer3d}}       &MM21     &3D                    & 57.4          & 42.1          \\
InstanceRefer \footnotesize{\cite{yuan2021instancerefer}}       &ICCV21       &3D                & 48.0            & 38.8          \\
3DVG-Transfor. \footnotesize{\cite{zhao20213dvg}}     &ICCV21       &3D                      & 51.4          & 40.8          \\
FFL-3DOG \footnotesize{\cite{feng2021free}}     &ICCV21       &3D                              & -             & 41.7          \\
SAT \footnotesize{\cite{yang2021sat}}       &ICCV21       &3D+2D                                    & 57.9          & \colorl{49.2}    \\
3DReferTrans. \footnotesize{\cite{abdelreheem20223dreftransformer}}     &WACV22     &3D & 47.0            & 39.0            \\
LanguageRefer \footnotesize{\cite{roh2022languagerefer}}        &CoRL22       &3D                 & 56.0            & 43.9          \\
3D-SPS~\footnotesize{\cite{luo20223d}}        &CVPR22       &3D+2D                 & \colorl{62.6}            & \colorm{51.5}          \\
BUTD-DETR \dag  \footnotesize{\cite{jain2022bottom}}       &ECCV22       &3D                        & \colorm{65.6}    & 49.1          \\
LAR~\footnotesize{\cite{bakr2022look}}        &NeurIPS22     &3D                        &  59.6    & 48.9          \\
\textbf{EDA (Ours)}      & -     &3D                                                             & \colorh{\textbf{68.1}} & \colorh{\textbf{52.1}} \\ 
\bottomrule
\end{tabular}
\caption{Performance on SR3D/NR3D datasets by Acc@0.25IoU as the metric. The detailed results of EDA in four subsets are provided in the \textcolor{magenta}{Supplementary Material}. \dag~Reevaluated by parsed text labels (see supp. for more details).}
\label{tab:SR3D_result}
\vspace{-20pt}
\end{table}

\vspace{-5pt}
\subsubsection{Ablation studies}

~~~~\textbf{Loss ablation.}
The ablation of the position-aligned loss and the semantic-aligned loss is shown in Table~\ref{tab:Ablation_loss}. The performance of the semantic-aligned loss is marginally better because its contrastive loss not only shortens the distance between similar text-visual features but also enlarges the distance between dis-matched features (such as $\boldsymbol{t}_{auxi}$ is a negative term in Eq.~(\ref{eq:loss_semantic})). Whereas position-aligned loss only considers object-related components (as in Eq.~(\ref{eq:P_text})).
When both losses supervise together, the best accuracy is achieved, demonstrating that they can produce complementary performance.

\textbf{Dense components ablation.} 
To demonstrate our insight into dense alignment, we perform ablation analysis on the different decoupled text components, and the results are displayed in the \textbf{``Regular VG"} column in Table~\ref{tab:Ablation}.
Analysis: 
\textbf{\romannumeral1)} (a) is our baseline implementation of the sparse concept, using only the ``Main object" component decoupled from the text. In contrast to BUTD-DETR, the text decoupling module (Sec.~\ref{sec:text}) is used to obtain text labels and features during training and inference instead of employing ground truth labels.
\textbf{\romannumeral2)} Dense-aligned sub-methods (b)-(h) outperform the sparse alignment (a) because of the finer-grained visual-linguistic feature fusion.
\textbf{\romannumeral3)} (b)-(e) indicate that adding any other component on top of the ``Main object" improves performance, demonstrating the validity of each text component. The ``Attribute" component aids in identifying characteristics such as colour, and shape, frequently mentioned in language descriptions. Unexpectedly, the ``Pronoun" component such as \textit{``it, that, and which"} have little meaning when used alone but also function in our method, indicating that the pronoun learned contextual information from the sentence. The ``Relationship" component facilitates comprehension of spatial relationships between objects. The component ``Auxiliary object" is a negative term in the loss (Eq.~(\ref{eq:loss_semantic})). During inference (Eq.~(\ref{eq:score})), its similarity is subtracted in the hopes that the predicted main object is as dissimilar to it as possible.
\textbf{\romannumeral4)} (f)-(h) integrate different components to make performance gains and reach the peak when all are involved, demonstrating that the functions of each component can be complementary, and there may be no overlap between the features of each one. The result reveals that our method effectively decouples and matches fine-grained multimodal features.

\begin{figure*}[!t]
\setlength{\abovecaptionskip}{2pt}
\centering
\includegraphics[width=1.0 \textwidth]{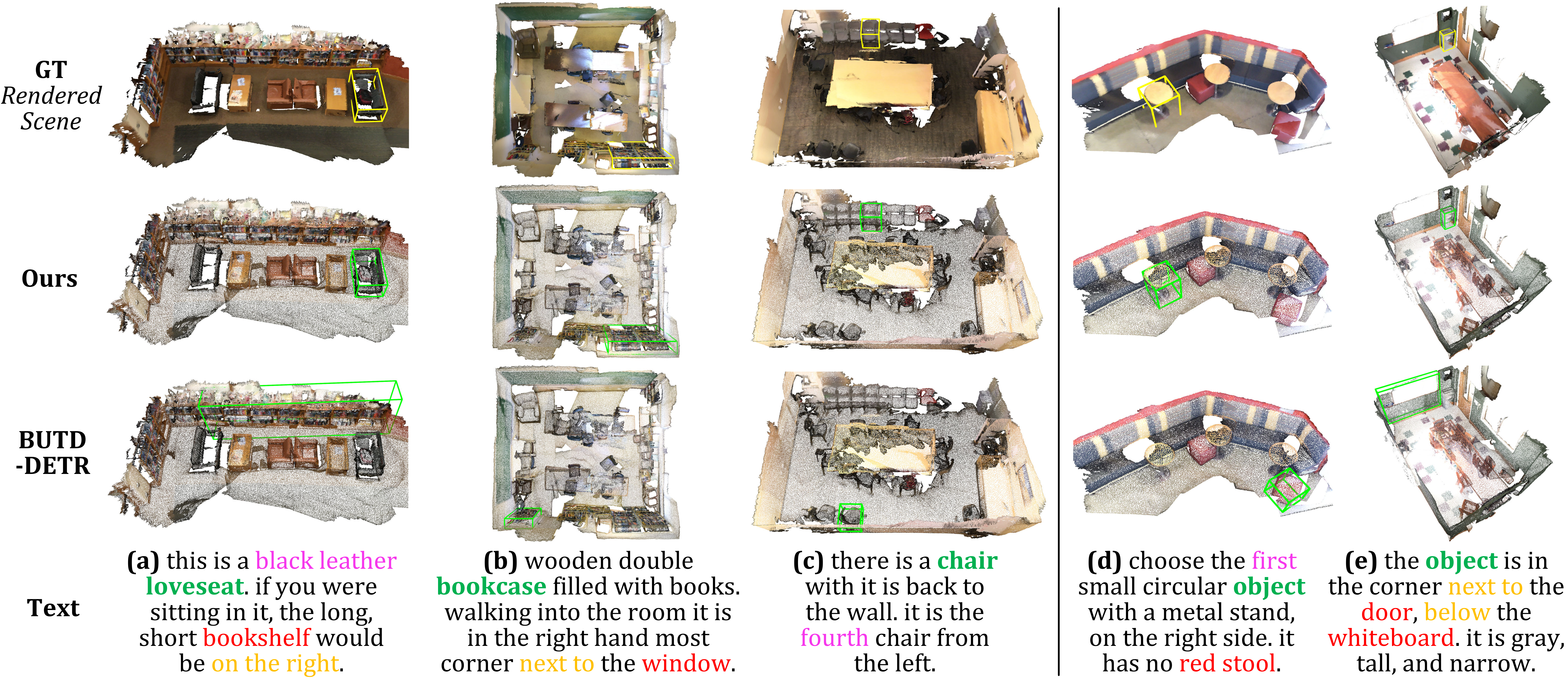} 
\caption{Qualitative results with ScanRefer texts. (a-c): Regular 3D visual grounding. (d-e): Grounding without object name.}
\label{fig:vis}
\vspace*{-10pt}
\end{figure*}

\begin{table}[!t]
\small
\centering
\renewcommand{\arraystretch}{1.0}
\setlength{\abovecaptionskip}{2pt}
\setlength{\tabcolsep}{8pt}
\begin{tabular}{c|cc|cc}
\toprule
\rowcolor[HTML]{E6E6E6} \multicolumn{1}{l|}{} & $+\mathcal{L}_{pos}$ & $+ \mathcal{L}_{sem}$ & Acc@0.25      & Acc@0.5       \\ 
\midrule
(a)                   & \ding{51}         &           & 51.2          & 39.6          \\
(b)                   &           & \ding{51}         & 52.2          & 39.9          \\
(c)                   & \ding{51}         & \ding{51}         & \textbf{54.6} & \textbf{42.3} \\ 
\bottomrule
\end{tabular}
\caption{Loss ablation on the ScanRefer dataset.}
\label{tab:Ablation_loss}
\vspace{-10pt}
\end{table}

\begin{table}[!t]
\footnotesize
\centering
\setlength{\tabcolsep}{3.5pt}
\setlength{\abovecaptionskip}{2pt}
\begin{tabular}{c|ccccc|cc|cc}
\bottomrule
\rowcolor[HTML]{E6E6E6}\multirow{2}{*}{ } & \multicolumn{5}{c|}{Dense components}                                                             & \multicolumn{2}{c|}{Regular VG} & \multicolumn{2}{c}{VG-w/o-ON} \\
\rowcolor[HTML]{E6E6E6}                        & \scriptsize{Main} & \scriptsize{Attri.}               & \scriptsize{Pron.}                & \scriptsize{Auxi.}                & \scriptsize{Rel.}                  & @0.25         & @0.5         & @0.25      & @0.5       \\ 
\midrule
(a)                     & \ding{51}    &                      &                      &                      &                       & 51.5             & 38.8            & 21.8          & 16.7          \\ 
\midrule
(b)                     & \ding{51}    & \ding{51}                    &                      &                      &                       & 53.1             & 41.3            & 23.2          & 17.8          \\
(c)                     & \ding{51}    &                      & \ding{51}                    &                      &                       & 52.9             & 40.8            & 23.7          & 18.6          \\
(d)                     & \ding{51}    & \multicolumn{1}{l}{} & \multicolumn{1}{l}{} & \ding{51}                    & \multicolumn{1}{l|}{} & 52.9             & 41.0            & 23.2          & 18.4          \\
(e)                     & \ding{51}    & \multicolumn{1}{l}{} & \multicolumn{1}{l}{} & \multicolumn{1}{l}{} & \ding{51}                     & 52.8             & 40.7            & 25.7          & 19.6          \\ 
\midrule
(f)                     & \ding{51}    & \ding{51}                    & \ding{51}                    &                      &                       & 53.8             & 41.8            & 24.2          & 18.5          \\
(g)                     & \ding{51}    & \ding{51}                    & \ding{51}                    & \ding{51}                    &                       & 54.2             & 42.3            & 24.0            & 18.8          \\
(h)                     & \ding{51}    & \ding{51}                    & \ding{51}                    & \ding{51}                    & \ding{51}                     & \textbf{54.6}    & \textbf{42.3}   & \textbf{26.5} & \textbf{21.2} \\ 
\bottomrule
\end{tabular}
\caption{Ablation studies of different text components on Regular VG and VG-w/o-ON tasks. Evaluated on the ScanRefer dataset.}
\label{tab:Ablation}
\vspace{-10pt}
\end{table}

%%%%%%%%% 4.2 %%%%%%%%%%%%%
\subsection{Grounding without Object Name (VG-w/o-ON)}
\label{sec:VG-w/o-ON}
\subsubsection{Experiment settings}
To evaluate the comprehensive reasoning ability of the model and avoid inductive biases about object names, we propose a new and more challenging task: grounding objects without mentioning object names (VG-w/o-ON). Specifically, we manually replace the object's name with \textit{``object"} in the ScanRefer validation set. For instance: \textit{``This is a brown wooden chair"} becomes \textit{``this is a brown wooden object"}. A total of 9253 samples were annotated and discarded another 255 ambiguous samples. 
We divide this language set into four subsets: only mentioning object attributes ($\sim$15\%), only mentioning spatial relationships ($\sim$20\%), mentioning both attributes and relationships ($\sim$63\%), and others ($\sim$2\%), as the first row in Table~\ref{tab:new_task}.
Note that, \textbf{without retraining}, we perform comparisons using our best model and comparative approaches' best models trained for the regular VG task (Sec. \ref{sec:RVG}).
\vspace{-6pt}

\subsubsection{Result and analysis}
Table~\ref{tab:new_task} reports the experimental results. 
\textbf{\romannumeral1)} The performance of all methods on this challenging task is significantly lower than that of regular VG (see Table.~\ref{tab:ScanRefer_result}), indicating that object categories provide rich semantic information, which is also conducive to classifying features. It's accessible to produce the category's inductive bias, especially in the ``unique" setting.
\textbf{\romannumeral2)} Our method achieves absolute lead with the overall performance of 26.5\% and 21.2\%, which is over 10\% higher than other methods. This preponderance demonstrates that our proposed text decoupling and dense alignment enable fine-grained visual-linguistic feature matching, where the model identifies the visual features most similar to other text components (\textit{e.g.} attributes, relations, \textit{etc.}).
\textbf{\romannumeral3)} Notably, in subset ``Attri+Rel", our method performs better than in the other subsets because additional cues can be exploited for fine-grained localization. However, the performance of the comparison approaches on this subset drops, revealing that more clues make them suffer from ambiguity.
\textbf{\romannumeral4)} Column ``VG-w/o-ON" in Table~\ref{tab:Ablation} shows ablation studies of text components for this new task. The performance increase offered by the additional components is more remarkable than during the regular VG task. Among them, the ``Relationship" component plays the most significant role because the spatial relationship with other objects may provide more obvious indications in this setting.
\textbf{\romannumeral5)} Fig.~\ref{fig:vis}(d-e) shows qualitative examples. Even without knowing the target's name, our method can infer it by other cues, while BUTD-DETR's performance drops catastrophically.

\begin{table}[!t]
\footnotesize
\centering
\setlength{\tabcolsep}{4pt}
\setlength{\abovecaptionskip}{2pt}
\begin{tabular}{c|ccc|cc}
\toprule
\rowcolor[HTML]{E6E6E6} & \multicolumn{3}{c|}{Subsets}      & \multicolumn{2}{c}{\textbf{Overall}} \\
\rowcolor[HTML]{E6E6E6} \multirow{-2}{*}{Method}                       & \scriptsize{Attri only}          & \scriptsize{Rel only}          & \scriptsize{Attri+Rel}          & \textbf{@0.25}    & \textbf{@0.5}    \\ 
\midrule
ScanRefer~\cite{chen2020scanrefer}               & 11.17          & 10.53           & 10.29           & 10.51              & 6.20             \\
TGNN~\cite{huang2021text}                    & 10.52          & 13.32          & 11.35          & 11.64             & \colorl{9.51}             \\
InstanceRefer~\cite{yuan2021instancerefer}           & 14.74          & 13.71          & 13.81          & \colorm{13.92}             & \colorm{11.47}            \\
BUTD-DETR~\cite{jain2022bottom}               & 12.30          & 12.11          & 11.86          & \colorl{11.99}             & 8.95             \\
\textbf{EDA (Ours)}           & \textbf{25.40} & \textbf{25.82} & \textbf{26.96} & \colorh{\textbf{26.50}}    & \colorh{\textbf{21.20}}   \\ 
\bottomrule
\end{tabular}
\caption{Performance of grounding without object name. The accuracy of subsets is measured by acc@0.25IoU, where the ``other" subset is not reported due to its small proportion.}
\label{tab:new_task}
\vspace*{-10pt}
\end{table}

%%%%%%%%%%%%%%%%%%%%%%%%%%%%%%%%%%%%%%%%%%%%%%%%%%%%%%%
% 5. Conclusion
%%%%%%%%%%%%%%%%%%%%%%%%%%%%%%%%%%%%%%%%%%%%%%%%%%%%%%%
\section{Conclusion}
We present EDA, an explicit, dense-aligned method for 3D Visual Grounding tasks. By decoupling text into multiple semantic components and densely aligning it with visual features under the supervision of position-aligned and semantic-aligned loss, we enable fine-grained visual-text feature fusion, avoiding the imbalance and ambiguity of existing methods. Extensive experiments and ablation demonstrate the superiority of our method. In addition, we propose a new challenging 3D VG subtask of grounding without the object name to evaluate model robustness comprehensively.
However, our limitations are the performance bottlenecks of multiple modules, such as PointNet++, RoBERTa, and text parsing modules. Especially when the text is lengthy, text decoupling may fail, resulting in performance degradation.

%%%%%%%%%%%%%%%%%%%%%%%%%%%%%%%%%%%%%%%%%%%%%%%%%%%%%%%
% 6. REFERENCES
%%%%%%%%%%%%%%%%%%%%%%%%%%%%%%%%%%%%%%%%%%%%%%%%%%%%%%%
\onecolumn
\twocolumn
{\small
\bibliographystyle{ieee_fullname}
\bibliography{egbib}
}

%%%%%%%%%%%%%%%%%%%%%%%%%%%%%%%%%%%%%%%%%%%%%%%%%%%%%%%
% 7. Supplementary Materials
%%%%%%%%%%%%%%%%%%%%%%%%%%%%%%%%%%%%%%%%%%%%%%%%%%%%%%%
\clearpage
\newpage
\label{page:supp}
\renewcommand\thesection{\Alph{section}}
\renewcommand\thesubsection{\thesection.\arabic{subsection}}
%\section*{\centering \Large Supplementary Materials}
\setcounter{section}{0}
\input{supp}
\clearpage

\end{document}

%% file: supp.tex
%%%%%%%%%%%%%%%%%%%%%%%%%%%%%%%%%%%%%%%%%%%%%%%%%%%%%%
\clearpage
\newcounter{counter}[section]
\twocolumn[{%
\renewcommand\twocolumn[1][]{#1}%
\begin{center}
    \Large
    \textbf{Supplementary Material for \\ EDA: Explicit Text-Decoupling and Dense Alignment for 3D Visual Grounding}
    \\[20pt]

    % author info
    \large
    Yanmin Wu\textsuperscript{\rm 1} \quad ~
    Xinhua Cheng\textsuperscript{\rm 1} \quad ~
    Renrui Zhang\textsuperscript{\rm 2,3} \quad ~
    Zesen Cheng\textsuperscript{\rm 1} \quad ~
    Jian Zhang\textsuperscript{\rm 1}$^*$
    \vspace{5pt}
    \\
    \textsuperscript{\rm 1} Shenzhen Graduate School, Peking University, China \\
    \textsuperscript{\rm 2} The Chinese University of Hong Kong, China  ~%\quad
    %\textsuperscript{\rm 3} Peng Cheng Laboratory \quad
    \textsuperscript{\rm 3} Shanghai AI Laboratory, China
    \\
    {\tt\small wuyanmin@stu.pku.edu.cn} \quad \quad
    {\tt\small zhangjian.sz@pku.edu.cn}
\end{center}
}]

Section~\textcolor{red}{A} of the supplementary material provides the implementation details of the individual modules and the network training details. In Section~\textcolor{red}{B} , we supplement with additional experiments and quantitative analyses. Finally, in Section~\textcolor{red}{C} , we present visualization results and qualitative analysis.

%%%%%%%%%%%%%%%%%%%%%%%%%%%%%%%%%
% 1. Implementation details
%%%%%%%%%%%%%%%%%%%%%%%%%%%%%%%%%
\section{Implementation details}
\label{sec:supp_1}
\textbf{Text decoupling module.}
The maximum length of the text is $l$$=$$256$, and the absence bit of the position label $L \in \mathbb{R}^{1 \times l}$ is padded with 0. 
Not every sentence can be decoupled into five semantic components, but the most fundamental ``main object" is required. 

\textbf{Encoder-Decoder.}
We keep hyperparameters consistent with BUTD-DETR~\cite{jain2022bottom}. The point cloud is tokenized as $\mathcal{V} \in \mathbb{R}^{n \times d}$ by the PointNet++~\cite{qi2017pointnet++} pre-trained on ScanNet. The text is tokenized as $\mathcal{T} \in \mathbb{R}^{l \times d}$ by the pre-trained RoBERTa~\cite{liu2019roberta}. Following object detection, the position and category of the boxes are embedded separately and concatenated as the box token $\mathcal{B} \in \mathbb{R}^{b \times d}$. The encoder, for visual-text feature extraction and modulation, is $N_{E}$$=$$3$ layers. The decoder with $N_{D}$$=$$6$ layers generates candidate object features $\mathcal{Q} \in \mathbb{R}^{k \times d}$. Where $n$$=$$1024$ denotes the number of seed points, $l$$=$$256$ the number of texts, $
b$$=$$132$ the number of detection boxes, $k$$=$$256$ the number of candidate objects, and $d$$=$$288$ the feature dimension. Please refer to BUTD-DETR for more details.

\textbf{Losses.}
\textbf{1)} In the \textbf{position-aligned loss} $\mathcal{L}_{pos}$, the weights of each component in Eq.~(1) are as follows: $\lambda_1$$=$$0.6$, $\lambda_2$$=$$\lambda_3$$=$$0.2$, $\lambda_4$$=$$0.1$.
These values indicate that the weight of the ``main object" component $L_{main}$ is higher, which is obvious. The ``relational" component $L_{rel}$ with lower weight because it affects both the main and auxiliary objects. See Sec.~\textcolor{red}{B.1.(4)} for parameter searching.
\textbf{2)} In the \textbf{semantic-aligned loss} $\mathcal{L}_{sem}$, the weight $w_{+}$ follows a similar trend. The four features $\boldsymbol{t}_{main}, \boldsymbol{t}_{attri}, \boldsymbol{t}_{pron}, \boldsymbol{t}_{rel}$ are weighted by $1.0, 0.2, 0.2$, and $0.1$, respectively. The weight $w_{-}$ acts on the negative item, where the feature weight of the auxiliary object is 2 and the remainder weighs 1. The purpose is to differentiate the features of the main object from the auxiliary objects. 
\textbf{3)} We optimize the model with the following \textbf{total loss}:
\begin{equation}
    \mathcal{L} = (\alpha(\mathcal{L}_{pos} + \mathcal{L}_{sem}) + 5\mathcal{L}_{box} + \mathcal{L}_{iou})/(N_D+1) + 8\mathcal{L}_{pts}, \tag{8}
    \label{eq:loss_semantic_text}
\end{equation}
where $\mathcal{L}_{pos}$ and $\mathcal{L}_{sem}$ represent the visual-language alignment loss. $\mathcal{L}_{box}$ and $\mathcal{L}_{iou}$ indicate the object detection loss~\cite{liu2021group}, with $\mathcal{L}_{box}$ representing the L1 regression loss of the object's position and size and $\mathcal{L}_{iou}$ representing the object's 3D IoU loss
$N_D$ is the layer number of the Decoder. $\mathcal{L}_{pts}$ is the KPS point samping loss~\cite{liu2021group}.
$\alpha$ takes the value 1 in the SR3D/NR3D dataset and 0.5 in the ScanRefer dataset. Because the SR3D/NR3D dataset provides the bounding box of candidate objects, while the ScanRefer dataset requires detecting the bounding box, we give higher weights for the detection loss in the ScanRefer dataset.

\textbf{Training details.}
The code is implemented based on PyTorch. We set the batch size to 12 on four 24-GB NVIDIA-RTX-3090 GPUs. For ScanRefer, we use a $2$$e$$-$$3$ learning rate for the visual encoder and a $2$$e$$-$$4$ learning rate for all other layers. It takes about 15 minutes per epoch, and around epoch 60, the best model appears. The learning rates for SR3D are $1$$e$$-$$3$ and $1$$e$$-$$4$, 25 minutes per epoch, requiring around 45 epochs of training. The learning rates for NR3D are set at $1$$e$$-$$3$ and $1$$e$$-$$4$, 15 minutes per epoch, and around 180 epochs are trained.
Since SR3D is composed of brief machine-generated sentences, convergence is easier. ScanRefer and NR3D are comprised of human-annotated free-form complex descriptions, respectively, and require more training time.

%%%%%%%%%%%%%%%%%%%%%%%%%%%%%%%%%
% 2. Additional experiments
%%%%%%%%%%%%%%%%%%%%%%%%%%%%%%%%%
\section{Additional experiments}
\label{sec:supp_2}

\subsection{Regular 3D Visual Grounding}
\label{sec:supp_rvg}
\textbf{(1) The explanation of the BUTD-DETR's performance.} Given a sentence, such as \textit{``It is a brown chair with armrests and four legs . It is directly under a blackboard"}, our text decoupling module determines that \textit{``chair"} is the main object based on grammatical analysis and thus obtains the position label $L_{main} = 0000100...$. However, in the official implementation of BUTD-DETR, which requires an additional ground truth class for the target object, its input is: \textit{``\textless object name\textgreater ~ chair. \textless Description\textgreater ~ It is a brown chair ..."}. Then search for the position where the object name \textit{``chair"} appears in the sentence as a position label. This operation presents some problems: 
\begin{itemize}
    \item \textbf{\romannumeral1)} It is unfair to use GT labels during inference; 
    \item \textbf{\romannumeral2)} Descriptions may employ synonyms for the category \textit{``chair,"} such as \textit{``armchair, office-chair, and loveseat,"} leading to a failed search position label;
    \item \textbf{\romannumeral3)} Sometimes, the object name is not mentioned, such as when it is replaced by the word \textit{``object."} In the NR3D validation set, BUTD-DETR removed 800 such challenging samples, and about 5\% did not participate in the evaluation.
\end{itemize}
To be fair, we re-evaluate it using the position labels obtained by the proposed text decoupling module, as displayed in the second row in Tab.~\ref{tab:BUTD-DETR}.

\begin{table}[!h]
\renewcommand{\thetable}{6}
\centering
\small
\begin{tabular}{c|cc|c|c}
\toprule
\rowcolor[HTML]{E6E6E6}              & \multicolumn{2}{c|}{ScanRefer} &  &  \\
\rowcolor[HTML]{E6E6E6}              & 0.25IoU           & 0.5IoU           &  \multirow{-2}{*}{SR3D}                     &  \multirow{-2}{*}{NR3D}                     \\ 
\midrule
Official      & 52.2           & 39.8          & 67.1                  & 55.4                  \\
Re-evaluation & 50.4           & 38.6          & 65.6                  & 49.1                  \\ 
\bottomrule
\end{tabular}
\caption{Performance of BUTD-DETR, where SR3D/NR3D only use Acc@0.25IoU as the metric.}
\label{tab:BUTD-DETR}
\end{table}

\textbf{(2) Evaluation of ScanRefer using GT box.}
In the ScanRefer dataset, only the point cloud is provided as visual input, requiring object detection and language-based object grounding. Conversely, SR3D/NR3D offers additional GT boxes of candidate objects. Therefore, we further evaluate the ScanRefer dataset by GT boxes. As shown in Tab.~\ref{tab:gt-scanrefer}, our performance improves significantly without retraining, particularly in the unique setting where accuracy exceeds 90\%. This result demonstrates that more accurate object detection can further enhance our performance.

\begin{table}[!h]
\renewcommand{\thetable}{7}
\centering
\small
\setlength{\tabcolsep}{5pt}
\begin{tabular}{c|cccc|cc}
\toprule
\rowcolor[HTML]{E6E6E6} & \multicolumn{2}{c}{Unique}      & \multicolumn{2}{c|}{Multiple}   & \multicolumn{2}{c}{\textbf{Overall}}   \\
\rowcolor[HTML]{E6E6E6}       \multirow{-2}{*}{Method}                 & 0.25           & 0.5            & 0.25           & 0.5            & \textbf{0.25}          & \textbf{0.5}           \\ 
\midrule
BUTD-DETR             & 85.62          & 68.64          & 46.07          & 35.51          & 52.0            & 40.5          \\
EDA (Ours)              & \textbf{90.91} & \textbf{75.33} & \textbf{51.71} & \textbf{40.66} & \textbf{57.6} & \textbf{45.8} \\ 
\bottomrule
\end{tabular}
\caption{Performance on ScanRefer using GT box. Our method presents significant advantages.}
\label{tab:gt-scanrefer}
\end{table}

\textbf{(3) Detailed results on the SR3D/NR3D dataset.} Due to page limitations, we only report overall performance in Table~2. Table~\ref{tab:SR3D/NR3D_detailed} breaks down the detailed results of our method into four subsets: easy, hard, view-dependent, and view-independent.

\begin{table}[!h]
\renewcommand{\thetable}{8}
\centering
\small
\setlength{\tabcolsep}{5pt}
\begin{tabular}{c|cccc|c}
\toprule
\rowcolor[HTML]{E6E6E6} Dataset & Easy & Hard & View-dep. & View-indep. & \textbf{Overall} \\
\midrule
SR3D    & 70.3 & 62.9 & 54.1      & 68.7        & \textbf{68.1}   \\
NR3D    & 58.2 & 46.1 & 50.2      & 53.1        & \textbf{52.1}    \\
\bottomrule
\end{tabular}
\caption{Detailed performances of our method on the SR3D/NR3D dataset with the metric of Acc@0.25IoU.}
\label{tab:SR3D/NR3D_detailed}
\end{table}

\textbf{(4) Parameter search for the weight $\lambda$.}
The representative results of a grid search on the weights in Eq.~(1) are presented in Table~\ref{tab:weight}.
Either of these options outperforms existing methods, demonstrating the efficiency of our dense alignment. As seen in (a), it is not optimal to treat all components equally because their functions are not equivalent. When giving $\lambda_1$ a higher weight (see (f, g)), it turns out that a weight that is too high would also lead to a decrease in performance, which may compromise the functionality of other components. $\lambda_1$ takes 0.6 as the best option, and the other items take 0.1 or 0.2. We select option (d) for implementation.

\begin{table}[!h]
\renewcommand{\thetable}{9}
\centering
\small
\setlength{\tabcolsep}{5pt}
\begin{tabular}{cc|cc}
\toprule
\rowcolor[HTML]{E6E6E6}    & $\lambda_1$, $\lambda_2$, $\lambda_3$, $\lambda_4$            & @0.25IoU & @0.5IoU \\ 
\midrule
(a) & 1.0, 1.0, 1.0, 1.0            & 53.6        & 40.6        \\
\rowcolor[HTML]{F5F5F5} (b) & 0.5, 0.2, 0.2, 0.2 & 53.3        & 40.9        \\
(c) & 0.6, 0.2, 0.2, 0.2 & 53.5        & 41.2        \\
\rowcolor[HTML]{F5F5F5} (d) & 0.6, 0.2, 0.2, 0.1 & \textbf{54.6}        & \textbf{42.3}        \\
(e) & 0.6, 0.1, 0.1, 0.1 & 54.5        & 42.0        \\
%\rowcolor[HTML]{F5F5F5} (f) & 0.7, 0.1, 0.1, 0.1 & 53.8        & 42.1        \\
\rowcolor[HTML]{F5F5F5} (f) & 0.8, 0.2, 0.2, 0.2 & 52.9        & 41.5        \\ 
(g) & 1.0, 0.2, 0.2, 0.2 & 53.2        & 40.3        \\ 
\bottomrule
\end{tabular}
\caption{Grid search of the weight $\lambda$. Evaluated on the ScanRefer dataset. We select (d) for implementation.}
\label{tab:weight}
\end{table}

\textbf{(5) Does the text component really help?}
Based on the ``main object", we densely align the other four text components (``Attribute", ``Relationship", ``Pronoun", and ``Auxiliary object") with visual features. The question immediately arises whether the random alignment with some words yields the same gain. As a comparison, we randomly select four words in the sentence to align with the visual features. As shown in Table~\ref{tab:random}(b), although the performance is improved by 0.5\%, it is lower than when only one of the four components is aligned (c,d) and substantially worse than aligning all four (e).
This 0.5\% gain may be due to the randomly aligned words with the possibility of involving four text components. The results demonstrate our insight into dense alignment and decoupling of meaningful components.

\begin{table}[!h]
\renewcommand{\thetable}{10}
\centering
\small
\resizebox{0.48\textwidth}{!}{
\setlength{\tabcolsep}{3pt}
\begin{tabular}{l|c|c|cc|c|c}
\toprule
\rowcolor[HTML]{E6E6E6}\multirow{2}{*}{} & \multirow{2}{*}{Main} &  & \multicolumn{2}{c|}{\makecell[c]{One of \\ \{\footnotesize{Attr, Pron, Auxi, Rel}\}}} &  & \multirow{2}{*}{Acc.} \\
\rowcolor[HTML]{E6E6E6}                  &                       &          \multirow{-3}{*}{\makecell[c]{Random \\ 4 words}}                       & ~~~~(lowest)                   & (best)                  &  \multirow{-3}{*}{\makecell[c]{\footnotesize{Attr + Pron +} \\ \footnotesize{Auxi + Rel}}}                                         &                       \\ 
\midrule
(a)               & \ding{51}                     &                                 &                            &                         &                                           & 51.5                  \\
\rowcolor[HTML]{F5F5F5} (b)               & \ding{51}                     & \ding{51}                               &                            &                         &                                           & 52.0                  \\
(c)               & \ding{51}                     &                                 & ~~~\ding{51}                          &                         &                                           & 52.8                  \\
\rowcolor[HTML]{F5F5F5} (d)               & \ding{51}                     &                                 &                            & \ding{51}                       &                                           & 53.1                  \\
(e)               & \ding{51}                     &                                 &                            &                         & \ding{51}                                         & \textbf{54.6}         \\ 
\bottomrule
\end{tabular}
}
\caption{Comparison with the alignment of four random words. The metric is Acc@0.25IoU. (a): Baseline, only aligned with the ``Main Object" text component; (b): aligned with four random words; (c-d): aligned with one of our four decoupled components; (e): aligned with all four components.}
\label{tab:random}
\end{table}

%%%%%%%%%%%%%%%%%%%%%%%%%%%%%%%%
\subsection{Language Modulated 3D Object Detection}

We maintain the same experimental setup as BUTD-DETR to evaluate the performance of 3D object detection on ScanNet. The 18 classes in ScanNet are concatenation into a sentence: \textit{``bed. bookshelf. cabinet. chair. counter. curtain. desk. door ..."} as text input. Output the bounding boxes and classes of all objects in the point cloud. As shown in Tab. \ref{tab:obj_det}, our model after text modulation on the ScanRefer achieves 1.1\% and 1.5\% higher performance than BUTD-DETR, to 64.1\% and 45.3\%. 
Note that the proposed method is not specifically designed for object detection, and the performance evaluation uses the \textbf{same model} as the visual grounding task (Table~1 and Table~5).

\begin{table}[!h]
\renewcommand{\thetable}{11}
\centering
\small
\begin{tabular}{c|p{1.3cm}<{\centering} p{1.3cm}<{\centering}}
\toprule
\rowcolor[HTML]{E6E6E6} Method                           & mAP@0.25      & mAP@0.5       \\ 
\midrule
DETR+KPS+iter \dag                   & 59.9          & -             \\
3DETR with PointNet++ \dag           & 61.7          & -             \\
BUTD-DETR \scriptsize{trained on ScanRefer} & 63.0          & 43.8          \\
EDA {\scriptsize{trained on ScanRefer}} (Ours)  & \textbf{64.1} & \textbf{45.3} \\ 
\bottomrule
\end{tabular}
\caption{3D Object detection performance on ScanNet. \dag~ The accuracy is provided by BUTD-DETR.}
\label{tab:obj_det}
\end{table}

%%%%%%%%%%%%%%%%%%%%%%%%%%%%%%%%%
% 3. Qualitative analysis
%%%%%%%%%%%%%%%%%%%%%%%%%%%%%%%%%
\section{Qualitative analysis}
\label{sec:supp_3}

\begin{figure*}[!t]
\centering
\includegraphics[width=1.0 \textwidth]{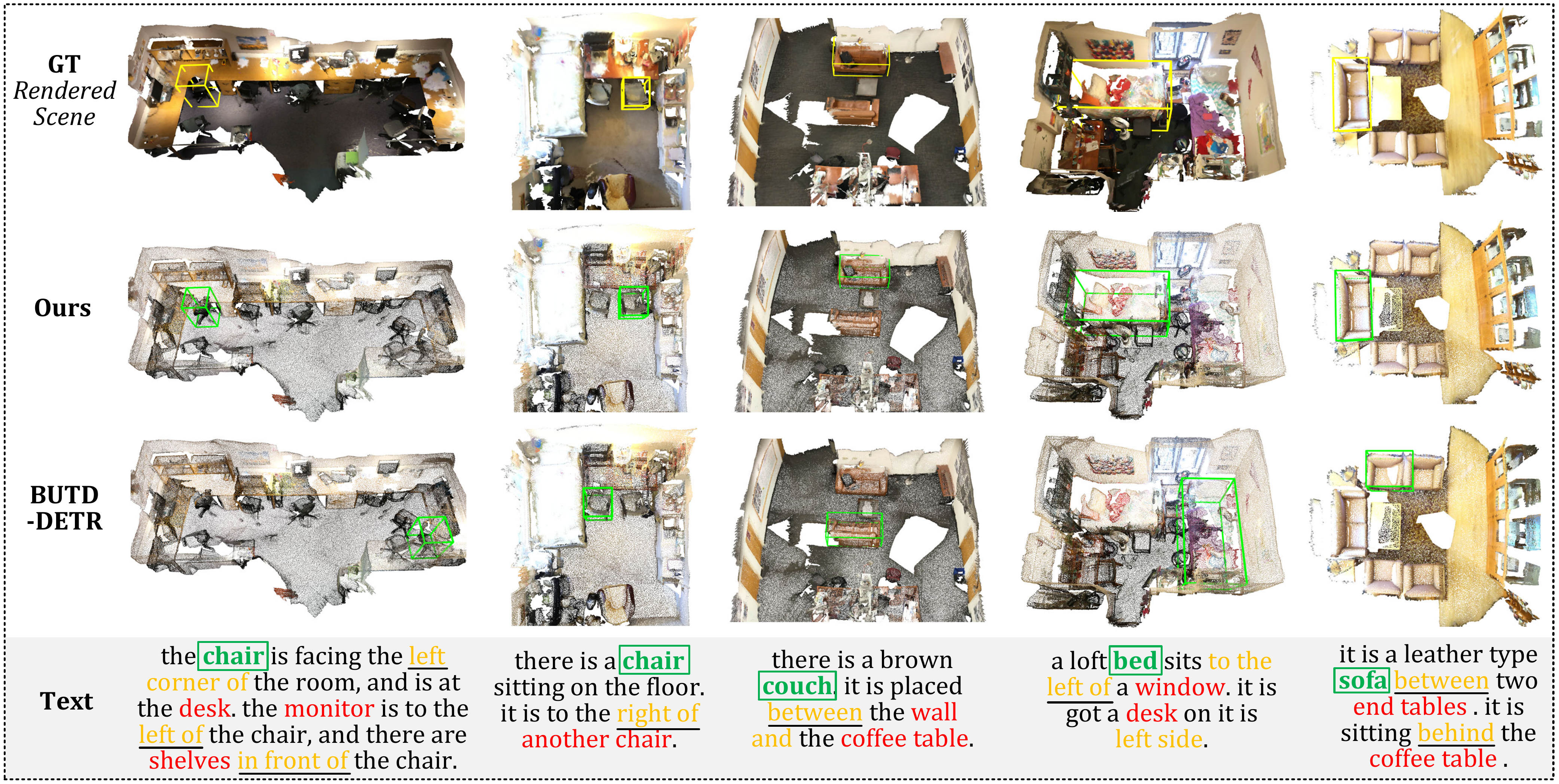} 
\caption{Qualitative comparison of the \textbf{regular 3D VG task}. Our method has a superior perception of \textbf{spatial relationships}.}
\label{fig:vis_rel}
\end{figure*}

\begin{figure*}[!t]
\centering
\includegraphics[width=1.0 \textwidth]{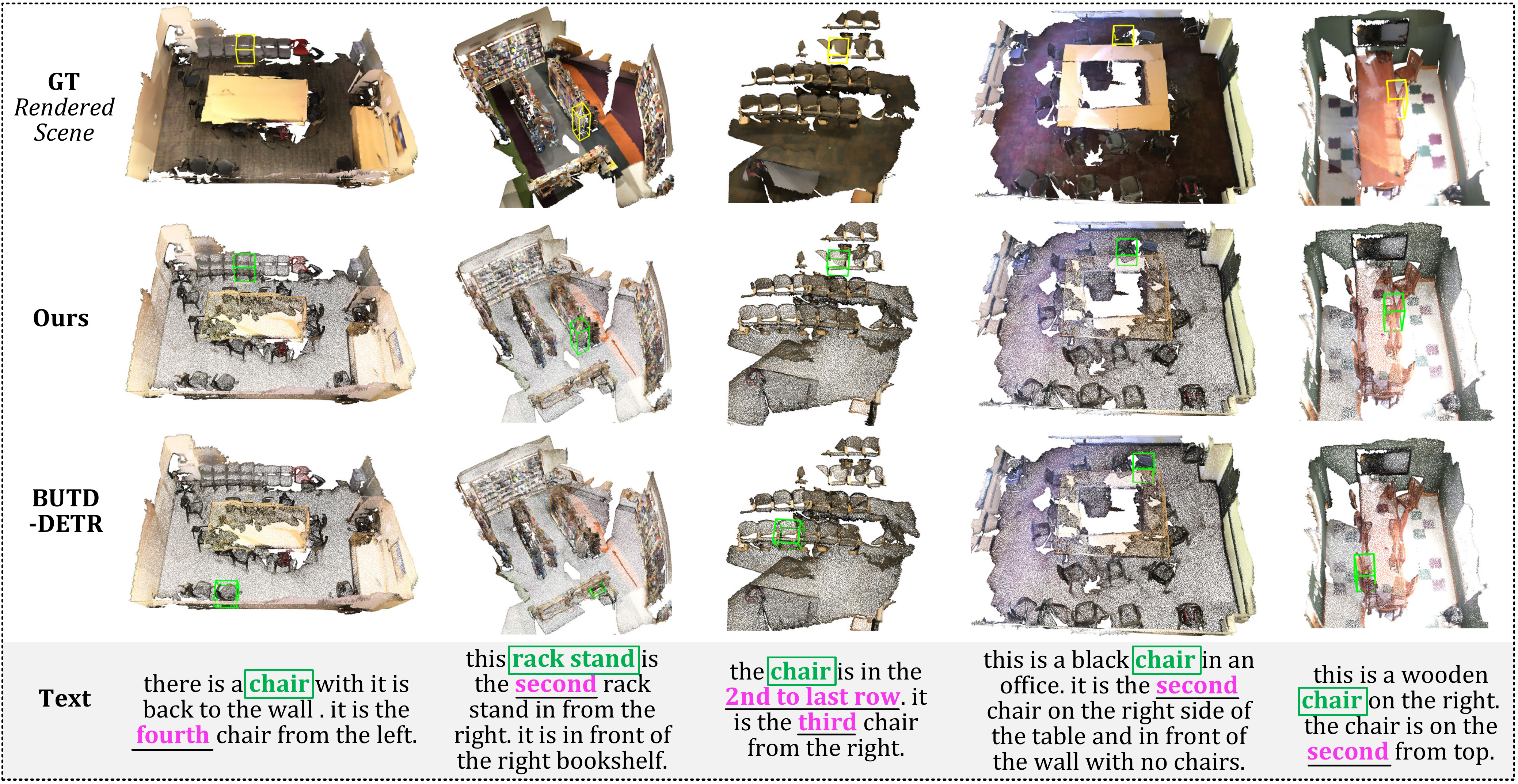} 
\caption{Qualitative comparison of the \textbf{regular 3D VG task}. Challenging cases with \textbf{ordinal numbers}.}
\label{fig:vis_num}
\end{figure*}

\textbf{1) Regular 3D Visual Grounding.}
Qualitative results on the regular 3D visual grounding task are displayed in Fig.~\ref{fig:vis_attri}, \ref{fig:vis_rel},  \ref{fig:vis_num}. 
\textbf{\romannumeral1)} Fig.~\ref{fig:vis_attri} indicates that compared to BUTD-DETR, our method has a superior perception of appearance, enabling the identification of objects based on their attributes among several candidates of the same class. This improvement is made possible by the alignment of our decoupled text attribute component with visual features.
\textbf{\romannumeral2)} Fig.~\ref{fig:vis_rel} demonstrates that our method exhibits excellent spatial awareness, such as orientation and position relationships between objects. The alignment of our decoupled relational component with visual features and the positional encoding of Transformer may be advantageous to this capability.
\textbf{\romannumeral3)} Furthermore, we surprisingly found that our method also has a solid understanding of ordinal numbers, as shown in Fig.~\ref{fig:vis_num}, probably because we parsed ordinal numbers as part of the attribute component of the object. These examples demonstrate that text decoupling and dense alignment enable fine-grained visual-linguistic matching.

\begin{figure*}[!t]
% \addtocounter{figure}{+4}
\centering
\includegraphics[width=1.0 \textwidth]{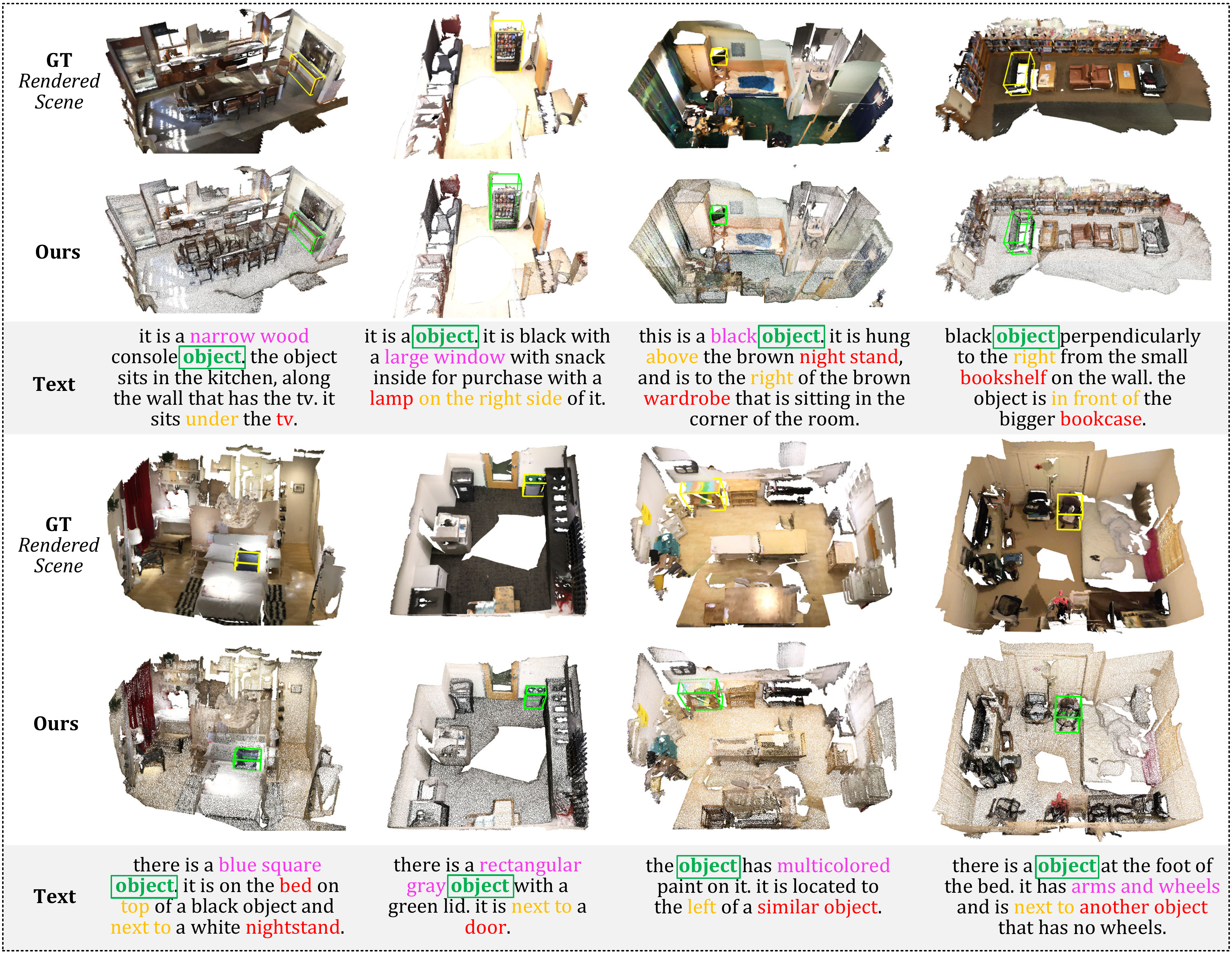} 
\caption{3D Visual grounding without object name (\textbf{VG-w/o-ON}), where the word ``object" replaces the target's name.}
\label{fig:vis_wo_on}
\end{figure*}

\textbf{2) Grounding without Object Name (VG-w/o-ON).}
The visualization results of this challenging task are shown in Fig.~\ref{fig:vis_wo_on}. Since the target object's name is not provided, the model must make inferences based on appearance and positional relationships with auxiliary objects. However, other contrastive methods perform weakly on this task because they rely heavily on object names to exclude interference candidates, weakening the learning of other attributes.

\begin{figure*}[]
\centering
\includegraphics[width=1.0 \textwidth]{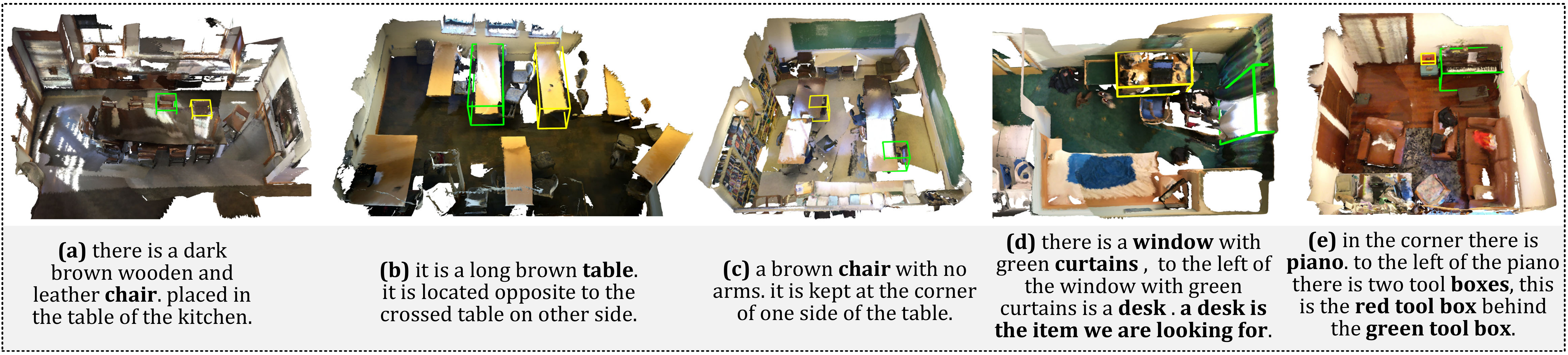} 
\caption{\textbf{Failure cases}, with the \setlength{\fboxrule}{1.0pt}\fcolorbox{yellow}{white}{\makebox(24,6){GT box}} and the \setlength{\fboxrule}{1.0pt}\fcolorbox{green}{white}{\makebox(48,6){predicted box}} shown in yellow and green, respectively. (a-c): Failure due to ambiguity of reference. (d-e): Failure due to text parsing error for complex and long sentences.}
\label{fig:vis_failure}
\end{figure*}

\textbf{3) Failure Case Analysis.} Although our method delivers state-of-the-art performance, there are still a significant number of failure occurrences, which we analyze visually. 
\textbf{\romannumeral1)} Many language descriptions are intrinsically ambiguous, as illustrated in Fig.~\ref{fig:vis_failure}(a-c), especially in the ``multiple" setting, the appearance attributes and spatial relationships of the target object are not unique, and there are multiple alternatives for candidate objects that match the requirements. 
\textbf{\romannumeral2)} The text parsing error may occur owing to the language description's complexity and diversity. Such as, the GT object in Fig.~\ref{fig:vis_failure}(d) is a desk, but we parse it as a window; the GT object in Fig.~\ref{fig:vis_failure}(e) is a box, but we parse it as a piano.
\textbf{\romannumeral3)} There are also some cases where cross-modal feature matching fails even though the text parses well. 

\begin{figure*}[!t]
\addtocounter{figure}{+4}
\centering
\includegraphics[width=1.0 \textwidth]{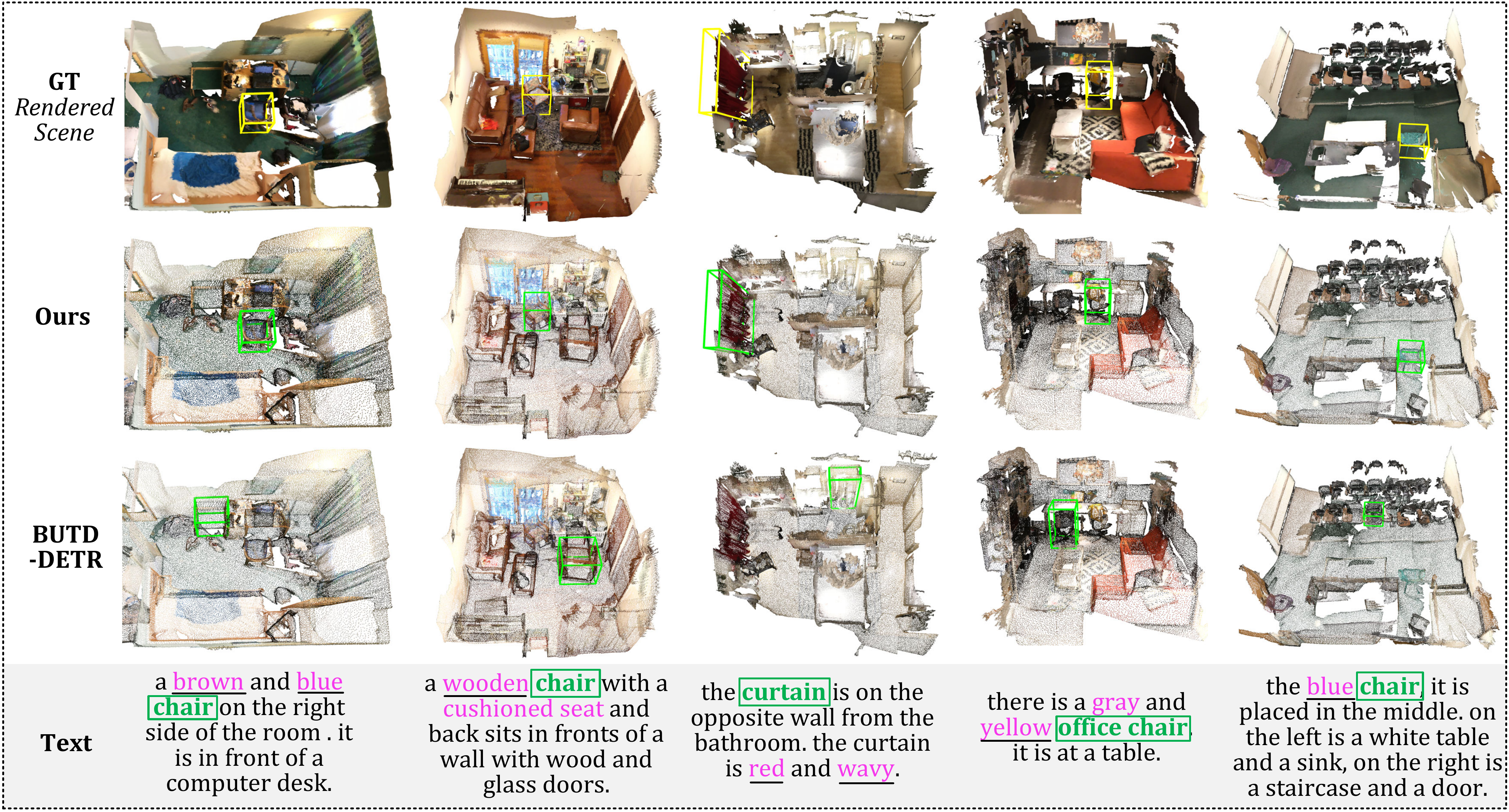} 
\caption{Qualitative comparison of the \textbf{regular 3D VG task}. Our method has a superior perception of \textbf{appearance attributes}.}
\label{fig:vis_attri}
\end{figure*}

% failure

% %%%%%%%%%%%%%%%%%%%%%%%%%%%%%%%%%%%%%%%%%%%%%%%%%%%%%%%
% % 6. REFERENCES
% %%%%%%%%%%%%%%%%%%%%%%%%%%%%%%%%%%%%%%%%%%%%%%%%%%%%%%%
% % \onecolumn
% % \twocolumn
% {\small
% \bibliographystyle{ieee_fullname}
% \bibliography{egbib}
% }